%% file: main.tex
\documentclass{clv3}

\usepackage{hyperref}
\usepackage{xcolor}
\usepackage{subcaption}
\usepackage{microtype}
\usepackage{array}
\usepackage{enumitem}
\usepackage{colortbl}
\usepackage{graphicx}
\usepackage{color}
\usepackage{soul}
\usepackage{verbatim}
\usepackage{multirow}
\usepackage{amsmath}
\usepackage{threeparttable} 
\usepackage{blindtext}
\usepackage{booktabs}
\usepackage{xpatch}
\usepackage{float}
\usepackage{placeins} 
\usepackage{flafter} 
\newcommand{\Revision}[1]{\textcolor{black}{#1}}
\definecolor{darkblue}{rgb}{0, 0, 0.5}
\hypersetup{colorlinks=true,citecolor=darkblue, linkcolor=darkblue, urlcolor=darkblue}

\begin{document}


\runningtitle{Computational Linguistics}

\runningauthor{Zheng, et al.}

\title{LMLPA: Language Model Linguistic Personality Assessment}

\author{Jingyao Zheng\thanks{Hung Hom, Kowloon, Hong Kong, 100872. E-mail: jingyao.zheng@connect.polyu.hk.}}
\affil{Hong Kong Polytechnic University}

\author{Xian Wang\thanks{Hung Hom, Kowloon, Hong Kong, 100872. E-mail: xiann.wang@connect.polyu.hk.}}
\affil{Hong Kong Polytechnic University}

\author{Simo Hosio\thanks{Pentti Kaiteran katu 1, Oulu, Finland, 90014. Email: simo.hosio@oulu.fi.}}
\affil{University of Oulu}

\author{Xiaoxian Xu\thanks{Trumpington St, Cambridge, United Kingdom, CB2 1PZ. Email:xx814@cam.ac.uk.}}
\affil{University of Cambridge}

\author{Lik-Hang Lee\thanks{Hung Hom, Kowloon, Hong Kong, 100872. E-mail: lik-hang.lee@polyu.edu.hk.}}
\affil{Hong Kong Polytechnic University}
\maketitle

\begin{abstract}
Large Language Models (LLMs) are increasingly used in everyday life and research. One of the most common use cases is conversational interactions, enabled by the language generation capabilities of LLMs. Just as between two humans, a conversation between an LLM-powered entity and a human depends on the personality of the conversants. However, measuring the personality of a given LLM is currently a challenge. This article introduces the Language Model Linguistic Personality Assessment (LMLPA), a system designed to evaluate the linguistic personalities of LLMs. Our system helps to understand LLMs' language generation capabilities by quantitatively assessing the distinct personality traits reflected in their linguistic outputs. Unlike traditional human-centric psychometrics, the LMLPA adapts a personality assessment questionnaire, specifically the Big Five Inventory, to align with the operational capabilities of LLMs, and also incorporates the findings from previous language-based personality measurement literature. To mitigate sensitivity to the order of options, our questionnaire is designed to be open-ended, resulting in textual answers. Thus, the Artificial Intelligence (AI) rater is needed to transform ambiguous personality information from text responses into clear numerical indicators of personality traits. Utilising Principal Component Analysis and reliability validations methods, our findings demonstrate that LLMs possess distinct personality traits that can be effectively quantified by the LMLPA. This research contributes to Human-Centered AI and Computational Linguistics, providing a robust framework for future studies to refine AI personality assessments and expand their applications in multiple areas, including education and manufacturing.
\end{abstract}

\input{Manuscript/Introduction}
\input{Manuscript/Related_work}
\input{Manuscript/Questionnaire}
\input{Manuscript/AIRaterAgent}

\input{Manuscript/WholeRatingSystem}
\input{Manuscript/Discussion}
\input{Manuscript/Conclusion}
\input{Manuscript/Appendix}

\newpage
\starttwocolumn
\bibliography{main}

\end{document}

%% file: Manuscript/Introduction.tex
\section{Introduction}
\begin{figure}[htbp]
    \centering
    \includegraphics[width=\linewidth]{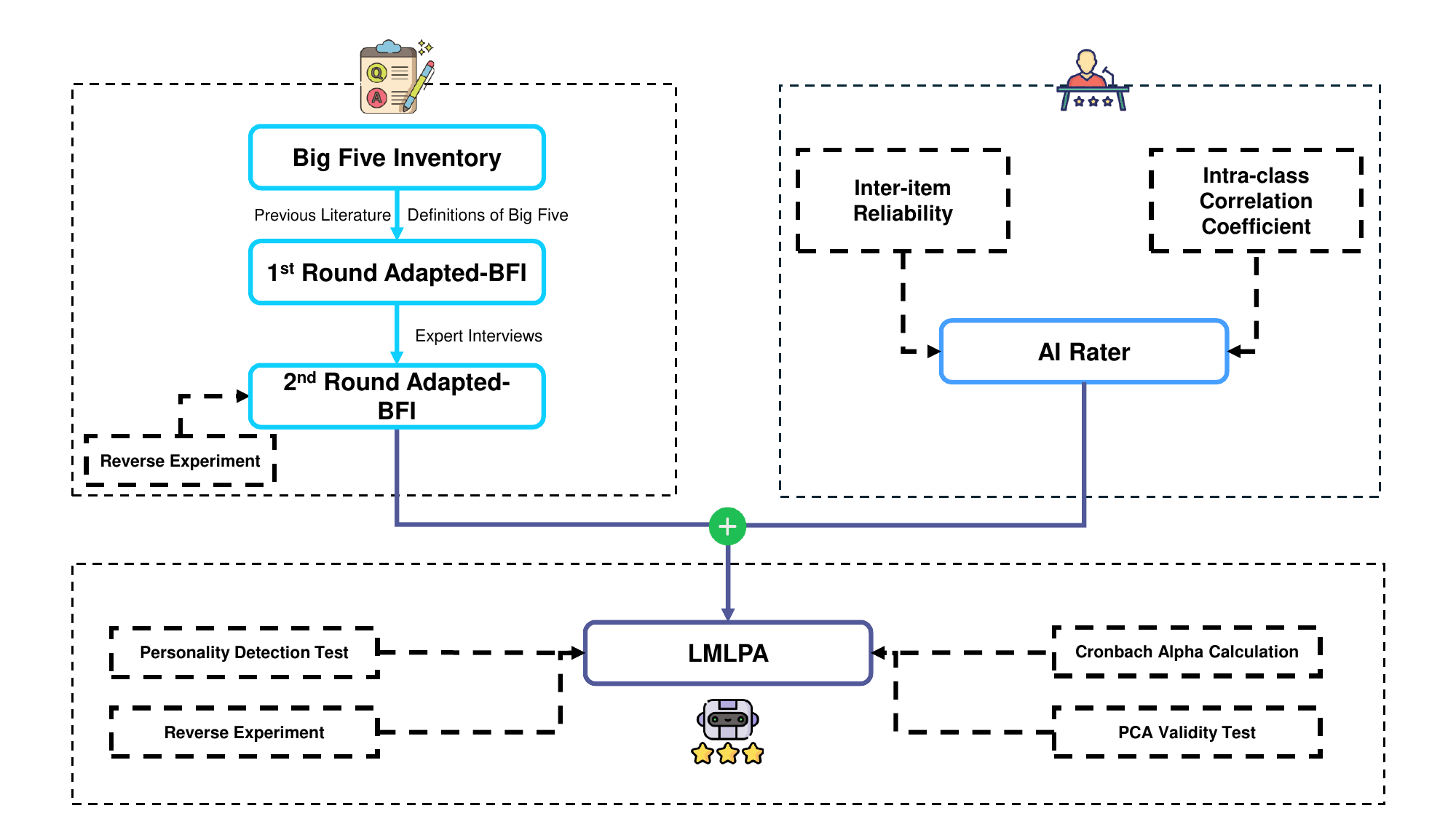}
    \caption{The workflow diagram illustrates the development process of LMLPA, an automated rating system designed to assess the personalities of Large Language Models (LLMs) based on their linguistic attributes. LMLPA comprises the Adapted Big Five Inventory and AI raters. The Adapted Big Five Inventory was developed by integrating the definitions of the Big Five personality traits with insights from previous language-based personality measurement literature. To ensure the reliability and validity of LMLPA, a series of rigorous reliability and validity tests were conducted, establishing the system's effectiveness in accurately assessing personality traits.}
    \label{fig:workflow}
\end{figure}
\Revision{The advent of Large Language Models (LLMs) has been a significant advancement in Artificial Intelligence (AI). LLMs have quickly demonstrated capabilities across diverse sectors such as education \cite{jeon2023large}, medicine \cite{thirunavukarasu2023large}, and learning analytics \cite{corlatescu2024automated}. A typical interaction affordance enabled by LLMs is conversation. Conversations depend on the personalities of the conversants. Therefore, better means to manage the personalities of LLM-powered interfaces is a step forward in facilitating further Human-Computer Interaction (HCI) research utilising LLMs and computational linguistic models. If we have greater control over the personality of a given LLM instance, or the conversational agent powered by the LLM, we can design better conversational interactions with LLM-powered systems. For instance, different types of recommendation engines can adapt to the human user's desired level of extroversion/introversion, thereby offering more engaging and compelling interactions. For example, in education, by assessing the personalities of LLMs, researchers can investigate which traits embodied by teaching AI could improve users' learning curves. They can instruct LLMs with different personalities, assess their personalities quantitatively, and conduct a user study to visualise the correlation between personality scores and students' performance and determine which personality trait is the most suitable for teaching AI \cite{sonlu2024effects}.}

\Revision{Multiple researchers believe that LLMs have personalities and have recognised the importance of exploring LLMs' personalities and have researched this area \cite{serapio2023personality, jiang2024evaluating, li2022does, pan2023llms}. The prevailing methodology often relies on self-reported questionnaires, originally designed for human personality assessment, to evaluate LLM personalities. For example, prior research on LLM personalities has predominantly employed questionnaires such as the Big Five Inventory (BFI) \cite{john1991big} and the International Personality Item Pool (IPIP) 50-item Big Five assessment \cite{goldberg2006international}, which are inappropriate for evaluating non-human entities like LLMs. Questions from these instruments, such as ``I see myself as someone who can be moody'' or ``I see myself as someone who can be cold and aloof'', presuppose emotional experiences that LLMs cannot possess. While these approaches are useful, they are also laborious and may arguably overlook fundamental differences between human beings and computers. Human personalities are complex and enduring constructs, encompassing traits, beliefs, emotions, and behaviours that define an individual's unique identity. In contrast, LLMs are algorithms trained on extensive textual data, tasked with understanding, predicting, and generating human language based on contextual input. They lack the cognitive and affective capacities inherent in human personalities. Thus, there is a pressing need to develop novel assessment tools tailored to evaluate LLM personalities in linguistic terms, aligning with their functional capabilities and operational paradigms. }

\Revision{Moreover, from a practical perspective, the self-reported answers are inconsistent with the option orders due to the LLMs' sensitivity to the Multiple-Choice Questions (MCQs). Many researchers, such as \citet{zheng2023large} and \citet{pezeshkpour2023large}, have provided evidence that the responses of LLMs are influenced by the order of options in MCQs. However, the self-reported questionnaires, like BFI and IPIP, consist of MCQs. Individuals need to indicate their level of agreement with the statement on a scale ranging from 1 to 5. Thus, administering these self-reported questionnaires leads to unreliable results. To prove this, \citet{gupta2023investigating} found that the same LLMs yield statistically different personality scores if the order of options in the questionnaires is modified, indicating the unreliability of this method. In our study (Section \ref{sub:whole-reverse}), we also conducted a similar reverse experiment by modifying the order of options and observing how many responses differed. We then calculated Cohen's Weighted Kappa values to assess the consistency of the results. Our findings, with a Cohen's Weighted Kappa of 0.401, support previous research \cite{gupta2023investigating}, indicating that the results are statistically inconsistent when the option order in the BFI is reversed. Consequently, the use of self-reported questionnaires for assessing the personalities of LLMs is arguable, highlighting a potential methodological flaw among the existing LLM personality assessment studies.}

\Revision{Considering the aforementioned limitations, we have developed a new framework that could automatically evaluate the LLMs' personalities quantitatively, only focusing on their language outputs. Contributing to computational linguistics, this framework, named the Language Model Linguistic Personality Assessment (LMLPA), is specifically designed to assess and characterise the personalities of LLMs by examining the linguistic patterns, style, and other language-related features present in their outputs. LMLPA establishes a foundational benchmark among inventories that assess linguistic personalities, particularly tailored to the unique characteristics of LLMs. \autoref{fig:workflow} illustrates the complete structure of our system, which consists of two main components: the Adapted-BFI and the AI Rater. Diverging from conventional self-assessment instruments, the Adapted-BFI adopts an open-ended format to accommodate the sensitivities LLMs exhibit towards the sequence and structure of multiple-choice options. The Adapted-BFI is based on the BFI, which is the foundation of multiple psychometric questionnaires, including BFI-2 \cite{soto2017next} and BFI-10 \cite{rammstedt2013short}, and incorporates insights from previous language-based personality measurement literature. Additionally, we conducted expert interviews to refine our questions, ensuring they accurately capture the intended personality traits through linguistic analysis. The adoption of open-ended questionnaires necessitates the integration of an AI agent capable of evaluating the responses. This AI agent will automate the scoring process, enabling systematic and consistent assessments of open-ended answers and facilitating the quantitative measurement of personality scores. In this study, we set the temperature of GPT-4-Turbo and Llama3 \cite{llama3modelcard} to 0 for all tests, ensuring the model's output is nearly deterministic. Temperature, in the context of language models like GPT, refers to a parameter that controls the level of randomness and diversity in generated text outputs, with higher values introducing more variability and lower values promoting more deterministic responses. By eliminating randomness in the generation process, we limit the test variables primarily to the effects of the specific test parameters being examined, such as the arrangement of frequency words.}

\Revision{This article contributes to a better understanding of the nuanced linguistic expressions of LLMs, through the development of the LMLPA. Utilising an Adapted-BFI and an AI Rater system, our framework evaluates personality traits through the lens of language usage, distinguishing it from traditional human-centric psychometrics. Another key advantage of this approach is its potential to mitigate the sensitivity of LLMs to the MCQs, thereby yielding statistically consistent results. A series of tests confirm the reliability and validity of the LMLPA, demonstrating its effectiveness in capturing distinct personality traits from LLM outputs. These results validate the framework's design and enhance our ability to predict and understand the behaviour of LLMs in diverse interaction scenarios. This research lays a foundation for further empirical studies, potentially leading to scientists being able to conduct studies that rely on different personalities of a conversational agent more easily.}

%% file: Manuscript/Related_work.tex
\section{Related Work}
Our work deals with personality theories, i.e. the Big Five (BF) model \cite{john1999big}, and their relevance to linguistic analysis.
This theoretical framework is essential for framing our study, as it highlights how established human personality assessment tools can be adapted for linguistic evaluation in LLMs.
Furthermore, our work focuses on LLM personality assessment. 
Here, we explain how traditional self-report questionnaires have been repurposed for LLMs and why new tools are needed.
\subsection{Personality Theories}
\begin{table}[htbp]
  \caption{This table illustrates the BF personality factors and their associated facets as defined by John and Srivastava \cite{john1999big}.}
    \renewcommand\arraystretch{1.20}
  \label{table:BF}
 \resizebox{1\linewidth}{!}{ 
  \begin{tabular}{p{100pt}p{400pt}}
    \toprule
    \textbf{Big Five Factor} &\textbf{Facet}\\
    \hline
    \rowcolor{gray!10}
    Openness & Ideas, Fantasy, Aesthetics, Actions, Feelings, Values\\
    Conscientiousness & Competence, Order, Dutifulness, Achievement striving, Self-discipline, Deliberation\\
    \rowcolor{gray!10} 
    Extraversion & Gregariousness, Assertiveness, Activity, Excitement-seeking, Positive emotions, Warmth\\
    Agreeableness & Trust, Straightforwardness, Altruism, Compliance, Modesty, Tender-mindedness\\
    \rowcolor{gray!10}
    Neuroticism & Anxiety, Angry hostility, Depression, Self-consciousness, Impulsiveness, Vulnerability\\
    \bottomrule
    \end{tabular}}
\end{table}
Personality is commonly defined as the consistent set of traits, attitudes, emotions, and behaviours that characterise individuals \cite{boyd2017language}. To explore these traits, psychologists have devised various personality theories, notably the BF \cite{john1999big} and the Myers-Briggs Type Indicator (MBTI) \cite{myers1985guide}. In our research, we utilise the BF as the framework to describe the personality dimensions of LLMs due to its widespread recognition and application across various fields, including clinical psychology \cite{nigg2002big, hilliard2014big}, industry \cite{castillo2017relationship, lounsbury2005investigation}, and education \cite{busato1998relation, smidt2015big}. \autoref{table:BF} lists the Big Five personality factors—Openness, Conscientiousness, Extraversion, Agreeableness, and Neuroticism—along with their specific facets as defined by \citet{john1999big}. However, the BF was originally developed to describe human behaviours \cite{digman1990personality} that covers emotions and other human features. To better suit the personality theories to the LLMs, its application in our study is confined to analysing linguistic patterns, styles, and other language-related features corresponding to these facets.

\subsection{Personality Measurement and Language}
These personality theories underpin the development of psychometric instruments. Instruments such as the Big Five Inventory (BFI) \cite{john1991big} and the NEO Personality Inventory-Revised (NEO-PI-R) \cite{costa2008revised} are widely used to assess the dimensions outlined by the BF model. Our questionnaire is adapted from the BFI, which serves as the foundation for many developed psychometric instruments. Various researchers have translated the BFI into different languages to facilitate broader usage and applicability \cite{alansari2016big, denissen2008development}. Historically, the BFI has been validated across various cultures and contexts, demonstrating its broad applicability and effectiveness in capturing a wide spectrum of personality traits \cite{hee2014validity, alansari2016big, worrell2004reliability}. Moreover, Soto and John \cite{soto2017next} expanded upon this by creating the Big Five Inventory-2 (BFI-2), which offers enhanced bandwidth, fidelity, and predictive power. Additionally, Rammstedt et al. condensed the BFI into a concise 10-item version, known as the BFI-10 \cite{rammstedt2013short}. Thus, its extensive validation and empirical reliability make it a robust model for BF personality trait assessment.

These popular instructions are self-report questionnaires that require participants to select scores based on their level of agreement with various statements. They are widely utilised in psychometrics due to their cost-effectiveness and the rapidity. However, the reliability of self-reported data is a subject of debate within the field of personality research. Critics argue that these self-assessment methods might not truly capture authentic personality traits but rather reflect respondents' subjective perceptions or explicit theories of their own characteristics \cite{mccrae1982self, morgeson2007reconsidering}. In response to these concerns, \citet{boyd2017language} highlighted that while self-report measures provide a subjective view of personality traits, language-based personality assessments offer an objective dimension. This approach builds on the historical belief that an individual's language use can unveil deeper psychological traits \cite{stone1966general}. Recent research has robustly demonstrated a significant correlation between personality traits and language use \cite{pennebaker1999linguistic, hirsh2009personality}, underscoring that numerous human behaviours are intricately encoded in language \cite{tausczik2010psychological}. A notable advancement in this field is the development of the Linguistic Inquiry and Word Count (LIWC) method \cite{pennebaker2001linguistic}. This methodological innovation enables the examination of the psychometric properties of language and facilitates the summarising of features from human texts, providing a sophisticated tool for linguistic analysis in psychological research. For example, \citet{jiang2023personallm}  utilised LIWC to explore the personalities of LLMs based on the context generated by LLMs.

To be more specific, we have drawn upon findings from existing research on language-based personality measurement to adjust the design of our questionnaire, details of which are presented in Section \ref{subsub:QuestionAdaption}. This integration not only strengthens our methodological approach but also aligns with our objective to systematically examine the nuances in language usage that characterise different personality dimensions. To the best of our knowledge, our inventory is the first open-ended questionnaire designed to measure the personalities of LLMs about their linguistic properties.
\subsection{Personalities of LLMs}
The study of personalities in LLMs has gained increasing importance as the LLMs are being widely utilised in multiple areas, including medicine \cite{thirunavukarasu2023large} and education \cite{kasneci2023chatgpt}. An increasing number of researchers have recognised the importance of LLMs' personality exploration in enhancing human-AI interaction. Among these researchers, most have applied self-report questionnaires, such as the BFI, which were originally designed for humans. Notable examples include the direct administration of the BFI by \citet{jiang2023personallm}, \citet{serapio2023personality}, \citet{pellert2023ai}, and \citet{li2022does} to assess LLM personalities. Beyond the BFI, the HEXACO series of questionnaires have also been widely employed. For instance, while Bodroza et al. \cite{bodroza2023personality} used the HEXACO-100 \cite{lee2018psychometric}, Miotto et al. \cite{miotto2022gpt} administered the HEXACO-60 \cite{ashton2009hexaco}.

Researchers have various methods for administering personality questionnaires to LLMs, including designing prompt instructions for LLMs to generate self-rated scores and employing a Zero-Shot Classifier (ZSC). Most researchers focus on crafting prompt instructions to obtain self-rated results effectively \cite{huang2023chatgpt, miotto2022gpt}, while a few utilised ZSC. \citet{pellert2023ai} used ZSC to determine the models' probability distribution across typical response options—ranging from ``disagree strongly'' to ``agree strongly''. These responses, corresponding to scores from 1 to 5, were then weighted by their probabilities to calculate expected values for each personality dimension. On the other hand, \citet{karra2022estimating} presented LLMs with the BFI and classified the resultant text responses. To be more specific, they utilised a Natural Language Inference (NLI) model to assess the accuracy of labels by setting the input text as the premise and creating hypotheses for each potential label. Building on \citet{karra2022estimating}'s methodology, we incorporated ZSC in Section \ref{sec:AI_Rater} to evaluate the efficacy and reliability of NLI-based ZSC as a tool for AI-based personality assessment.

However, it is noticeable that most researchers did not realise that psychometrics designed for human beings are unsuitable for LLMs, as LLMs lack emotions and actions. \citet{jiang2024evaluating} were pioneers in the field of adapting personality questionnaires for machines, marking a significant development in personality assessment methodologies. Their initiative involved modifying established questionnaires, specifically those derived from the IPIP and its IPIP-NEO derivatives \cite{goldberg2006international, goldberg1999broad, johnson2005ascertaining, johnson2014measuring}, along with the Big Five Inventory-Short (BFI-S) \cite{lang2011short}, to evaluate the BF personality traits in LLMs. Despite this approach's novelty, the researchers did not comprehensively detail the adaptation process. Importantly, the questionnaires retained a self-report format, which might not be entirely appropriate for LLMs due to inherent biases \cite{gupta2023investigating}. Moreover, the study lacked crucial reliability and validity assessments that are essential for ensuring that the adapted tools accurately and reliably measure personality traits in non-human entities. In contrast, while the research by \citet{serapio2023personality} stands out for conducting reliability and validity tests to affirm their results' consistency and validity, they similarly relied on direct administration of traditional self-report questionnaires. This approach may not fully capture the unique dynamics of personality expression in LLMs, thereby limiting the applicability of their findings in the context of artificial agents. Interestingly, \citet{serapio2023personality} employed a prompt template that integrates persona descriptions and personality descriptions to generate numerous responses for reliability and validity tests. Inspired by their methodology, we adopted a similar template; however, we incorporated these descriptions into the system prompt rather than the user prompt. By embedding the personality descriptions directly into the system prompt, we reduce the potential variability and influence of user-specific instructions, thereby enhancing the reliability and accuracy of our personality assessment framework.

These gaps underscore the need for further research to develop and validate assessment tools tailored to LLMs' capabilities and characteristics, ensuring that personality evaluations are scientifically robust and contextually appropriate. To address these issues, our research has developed a novel framework for evaluating the linguistic personalities of LLMs. It integrates both traditional psychometric principles and advanced AI methodologies. Furthermore, we have conducted a series of reliability and validity tests to demonstrate that our system effectively measures the LLMs' personalities.

%% file: Manuscript/Questionnaire.tex
\section{Questionnaire Development}
\label{sec:questionnaire}
Psychometric questionnaires are designed to measure human personalities, encapsulating emotions, actions, and behaviours. Therefore, it is essential to develop a questionnaire specifically targeting the linguistic properties of LLMs. This section delves into the development and validation of a novel questionnaire specifically designed to assess the linguistic personalities of LLMs. \Revision{To be more specific, we first adapted the traditional personality questions into formats suitable for LLMs, focusing on linguistic responses rather than human behaviours, such as emotional expressions (See Section \ref{subsub:QuestionAdaption}). Then, psychology experts were invited to validate our adaptation (See Section \ref{subsub:expert}). The design of the instruction prompt is also a crucial part of the questionnaire design since our targeted test takers are LLMs (See Section \ref{subsub: Instruction}). Notably, our hypothesis is that our questionnaire could avoid the LLMs' sensitivity to MCQs to provide more reliable results. Thus, Section \ref{sub:reverse} was conducted to demonstrate our hypothesis.}

\subsection{Questionnaire Design}
\subsubsection{Question Adaption}
\label{subsub:QuestionAdaption}
The adaptation of the BFI questions to the LMLPA inventory primarily involved reformulating the original items into open-ended questions that are better suited for assessing language use. Originally, the BFI items begin with ``I see myself as someone who...'', structured as MCQs for agreement. Respondents need to select from options among ``\textit{Strongly Disagree}'', ``\textit{Disagree}'', ``\textit{Neither Agree Nor Disagree}'', ``\textit{Agree}'', and ``\textit{Strongly Agree}''. This format can introduce the bias related to the order sensitivity of the options, which LLMs are particularly susceptible to \cite{gupta2023investigating}. To mitigate this bias, we converted them into open-ended items that begin with ``To what extent do you ...'', listed in \hyperref[apx:questions]{Appendix A}.

The further modification of questions, which aimed to make questions better suited for assessing language use, was guided by the foundation descriptions of BF personalities \cite{john1999big} along with prior research on linguistic analysis of personality \cite{pennebaker1999linguistic, boyd2017language}. Each BF personality trait question was reinterpreted to reflect observable language behaviours, considering the practical language capabilities of LLMs. Questions originally about human behaviours, such as work performance, thoughts, and emotions, were rephrased to correspond to linguistic attributes relevant to LLMs. For instance, the human-oriented question, ``I see myself as someone who does things efficiently.'' was adapted for LLMs to, ``To what extent do you utilise your training dataset to answer questions efficiently?''. This adaptation better suits the operational nature of LLMs, whose ``\textit{work}'' is limited to learning from training datasets and responding to user queries. Moreover, the question, ``I see myself as someone who has few artistic interests.'', was transformed to, ``To what extent do you exhibit a limited range or depth in generating responses related to artistic and creative topics?''. Unlike humans, LLMs do not possess personal interests in the conventional sense but can demonstrate varying levels of proficiency and depth in their output on specific subjects based on their training data. Such adaptations ensure that the questions effectively measure the intended traits by considering how LLMs generate and process text, thereby aligning the assessment criteria with the functional characteristics of these models. Similar methods have been implemented to adapt all original questions in BFI.

Building on insights from existing linguistic personality research, further adaptations were made to validate the adapted items. For example, the original BFI question ``I see myself as someone who is depressed, blue.'' was adapted based on findings from \citet{pennebaker1999linguistic}'s research, which suggests that a high neuroticism score correlates with frequent use of negative emotion words. Therefore, the question was reformulated to suit LLMs better: ``To what extent do you generate text expressing sadness, hopelessness, or low energy?''. Furthermore, two more examples are drawn from the Extraversion questions: ``I see myself as someone who generates a lot of enthusiasm.'' and ``I see myself as someone who has an assertive personality.''. Researchers \cite{pennebaker1999linguistic,lucas2001understanding, gill2019taking} have found positive correlations between the Extraversion scores and the use of positive emotion words. Also, \citet{gill2019taking}'s findings suggest that individuals with high Extraversion scores are more likely to use confident and assertive expressions. Thus, the adapted questions were framed as ``To what extent do you use exclamation points or express strong positive emotions?'' and ``To what extent do you tend to use confident and assertive words or expressions?''.
\subsubsection{Expert Interviews}\label{subsub:expert}
\paragraph{Participants}
\begin{table}[!ht]
  \caption{This table presents the demographics of the expert participants.}
  \label{table:Expert}
 \resizebox{1\linewidth}{!}{ 
  \begin{tabular}{p{80pt}p{180pt}p{120pt}p{160pt}}
    \toprule
    \textbf{Code} &\textbf{Years of Psychological Research Experience}&\textbf{Profession} & \textbf{Familiarity Ratings with BF}\\
    \hline
    \rowcolor{gray!10} 
    $P1$ &11& Psychological Counsellor & 3\\
    $P2$ &5& Psychological Researcher&5\\
    \rowcolor{gray!10} 
    $P3$ &4&  Psychological Researcher& 5  \\
    $P4$ &4&  Psychological Researcher& 5\\
    \rowcolor{gray!10}
    $P5$ &3.5&  Psychological Researcher & 2\\
    \bottomrule
\end{tabular}}
\end{table}

Following the initial development of the LMLPA inventory, we conducted expert interviews. These consultations aimed to draw on the expertise of psychologists to refine further and validate the questions. This step is crucial to ensure that the inventory accurately measures the intended personality dimensions through the language used by LLMs. This study has received approval from the University Institutional Research Board (IRB) (HSEARS-2024-0129002).

Prior to the interviews, we collected their basic demographics, including years of psychological research experiences and profession, and requested them to assess their knowledge of the BF traits using a 5-point Likert scale. A rating of 5 indicates very familiar and 1 indicates very unfamiliar. Based on their self-reported results, five experts were psychological researchers and counsellors, and had 3.5 to 11 years of experience in the psychology field (Mean = 5.5, and Standard Deviation = 3.12). Each expert also had practical experience with LLMs such as ChatGPT. Among the experts, three (\textit{P2}, \textit{P3}, and \textit{P4}) rated their familiarity as 5. It demonstrates a high level of understanding of BF personality traits, while one expert gave a rating of 3 (\textit{P1}) and another gave a rating of 2 (\textit{P5}) as shown in \autoref{table:Expert}. The mean familiarity rating across all experts is 4, with a standard deviation of 1.414. It reflects a generally high level of expertise regarding the BF personality traits. Despite \textit{P5}'s relatively low familiarity with the BF personality traits, her extensive expertise in questionnaire design proved highly beneficial. While other experts contributed more to ensure that our adapted questions remain true to the original definitions and facets of the BF, \textit{P5} played a crucial role in refining the questions to eliminate any ambiguity. Her adjustments were key in preventing potential irrelevant or misinterpreted responses by the LLM.
\paragraph{Procedure}
Each interview took around one hour and involved only one expert. For compensation, each expert received $13.3\$$ after the interview. We started with an introduction to the project's background and objectives. Our primary aim was to adapt the BFI for evaluating linguistic personalities based on the outputs of LLMs, thereby we provided experts with a copy of the original BFI for reference. Then, we presented the Adapted BFI questionnaire (see Section~\ref{subsub:QuestionAdaption} and \hyperref[apx:questions]{Appendix A}) and allocated 15-20 minutes for the experts to review the document thoroughly. During this period, we encouraged them to ask clarifying questions to fully understand the adaptations made and the rationale behind each modification.
\paragraph{Results} 
During the feedback session, almost all the experts agreed with the modifications proposed for the Openness and Extraversion dimensions. Experts \textit{P3} and \textit{P5} suggested only a few specific adjustments to enhance a few questions. For instance, \textit{P3} refined the question ``To what extent do you not strive to answer questions with self-motivation?'' to ``To what extent do you not strive to answer questions with elaborate responses?''. This change acknowledges that LLMs lack self-motivation, and redefines the focus to the quality of the response, emphasising that ``\textit{elaborate}'' implies a response that is ``carefully prepared and organised''. These targeted adjustments are instrumental in aligning the questions more closely with the capabilities and behaviours of LLMs, thereby improving the assessment's accuracy and relevance. On a similar note, \textit{P}5 offered a specific suggestion while adapting the BFI item that assesses Extraversion, particularly the traits of ``being outgoing, sociable''. She proposed enhancing the clarity and focus of the question by including the word ``\textit{informal}'' before ``\textit{response}''. This modification would refine the question to ``To what extent do you generate informal responses that facilitate interactive and engaging dialogue across diverse topics?''. Adding ``\textit{informal}'' aims to capture the casual and relaxed nature often associated with sociable and outgoing behaviours. It provides a more nuanced insight into how LLMs mimic human-like sociable interactions in their language output. This adjustment is consistent with the findings of \citet{gill2019taking}'s work that a high score in Extraversion is correlated with the frequent use of informal language.

However, there was a notable divergence in opinions between experts and us concerning the adaptations made to the Conscientiousness and Agreeableness dimensions, where several experts recommended more extensive changes. This feedback suggests that the initial adaptations may not fully capture the intended traits or could benefit from greater specificity on language to better align with the nuances of these personality aspects. Specifically, expert \textit{P2} proposed a significant revision to the question intended to assess Agreeableness. The original question, ``To what extent do you adapt to contradictory instructions or information that conflicts with previous knowledge?'' was rephrased to ``To what extent do you respond in a kind manner even if the user prompt is rude and offensive?''. This change was made to more accurately capture the essence of having a ``\textit{forgiving nature}'', focusing on the LLM's ability to maintain courteous interactions despite negative user prompts. In addressing the dimension of Conscientiousness, expert \textit{P4} pointed out that the original BFI question, which assesses tendencies toward being ``\textit{disorganized}'', traditionally pertains more to general lifestyle and work habits rather than specifically to language structure. Therefore, to better tailor this question for assessing LLMs, the focus was shifted towards the LLMs’ performance in generating coherent and logical responses. Consequently, the question was revised from ``To what extent do you generate disorganized sentences?'' to ``To what extent do you tend to generate non-logical answers?''. This adaptation more accurately captures the aspect of Conscientiousness that relates to orderliness and methodical thinking in the context of LLM output, emphasising the model's ability to maintain logical consistency in its responses.

Lastly, all experts highlighted the challenges associated with adapting questions related to Neuroticism, given the complex and subjective nature of internal emotional expressions and the stability that it entails. Especially for the questions that capture the changes in emotions, exemplified by the item ``I see myself as someone who gets nervous easily.', Experts \textit{P1}, \textit{P2}, and \textit{P3} supported our approach of incorporating stress-inducing situations into questions designed to assess the neuroticism trait. Specifically, \textit{P1} agreed with the approach, noting, ``It is beneficial to create a stressful scenario with conflict when measuring neuroticism''. \textit{P2} and \textit{P3} agreed that such challenging scenarios effectively elicit responses that reveal key aspects of linguistic stability and reactivity. 
\subsubsection{Instruction Prompt Design}
\label{subsub: Instruction}
\Revision{Since questions are designed for LLMs, the instruction prompt should be appropriately crafted to ensure that the LLMs can accurately and effectively respond to the questions. It ensures that LLMs comprehend the context and respond appropriately, mirroring traditional personality tests where participants are aware that they are testing their personalities.} \Revision{Thus, at the beginning of the prompt, we included a sentence, ``You are about to participate in a personality test.''. Furthermore,} to facilitate the rating, the prompts were specifically designed to instruct LLMs to include one of the following phrases: ``\textit{always}'', ``\textit{often}'', ``\textit{sometimes}'', ``\textit{rarely}'' and ``\textit{never}'' in their responses. These terms, which were utilised by \citet{caron2022identifying} to correspond to numerical scores from 1 to 5, do not serve as direct answer choices. Otherwise, our questions would become the MCQs again. The frequency words serve as reference points in our research that facilitate standardising the responses for consistent assessment. This was also agreed by experts. During Interview \ref{subsub:expert}, some experts stated including these frequency words simplifies the scoring process by providing references. \Revision{This consideration ended up with the following sentences: ``You will be given an open-ended
question.'' and ``Please carefully answer the question and contain phrases (always, often, sometimes,
rarely, never) in your answers.''.}

Additionally, further instruction was integrated that requires LLMs to provide a rationale for their choice of frequency phrase. \Revision{It corresponds to the sentence, ``Your response should be explained in a single and coherent sentence.'', in the prompt.} This requirement ensures that raters do not solely rely on the selected adverb of frequency but also consider the context and reasoning provided by the LLM. This dual-layered response mechanism significantly reduces the potential biases associated with traditional multiple-choice formats and improves the overall reliability and depth of the assessments. The \Revision{complete} template for these instruction prompts is detailed in \hyperref[apx:prompts]{Appendix B}. By requiring substantiated and contextually enriched responses, this method enhances the accuracy and interpretability of personality assessments conducted with LLM.

\subsection{Reverse Experiments and Results} 
\label{sub:reverse}
The primary aim of this experiment is to demonstrate the effectiveness of open-ended questionnaires and the reliability of instruction prompts in eliciting consistent and unbiased responses from LLMs. We investigated whether reversing the order of frequency words in the instruction prompt affects the responses generated by LLMs, using the widely recognised GPT-4-Turbo due to its extensive use in research and daily applications. The instruction prompt ``... contain one of the following phrases (always, often, sometimes, rarely, never) ...''  was reversed to `` ... contain one of the following phrases (never, rarely, sometimes, often, always) ...''. This methodology has been similarly employed in studies examining the effects of option-order or scale reversal in human self-assessment tests \cite{rammstedt2007does}, aiming to show that the outcomes of personality assessments remain consistent regardless of the sequence in which options are presented. To ensure GPT-4-Turbo would not be affected by the questions, each question was answered independently by calling the model's API separately.

To analyse the consistency, this study \Revision{examines the chosen keywords in the answers and} assesses the sentence semantic similarity between answers generated before and after the reversal of frequency word order. The 44 responses generated by GPT-4-Turbo were embedded using the sentence-transformers model ``all-MiniLM-L6-v2'' \footnote{https://huggingface.co/sentence-transformers/all-MiniLM-L6-v2}. This model builds upon the pretrained MiniLM architecture \cite{wang2020minilm} and has been further fine-tuned on a dataset comprising 1 billion sentence pairs, enhancing its capability to capture nuanced semantic meanings from texts.
\begin{figure}[h!]
    \centering
    \includegraphics[width=\linewidth]{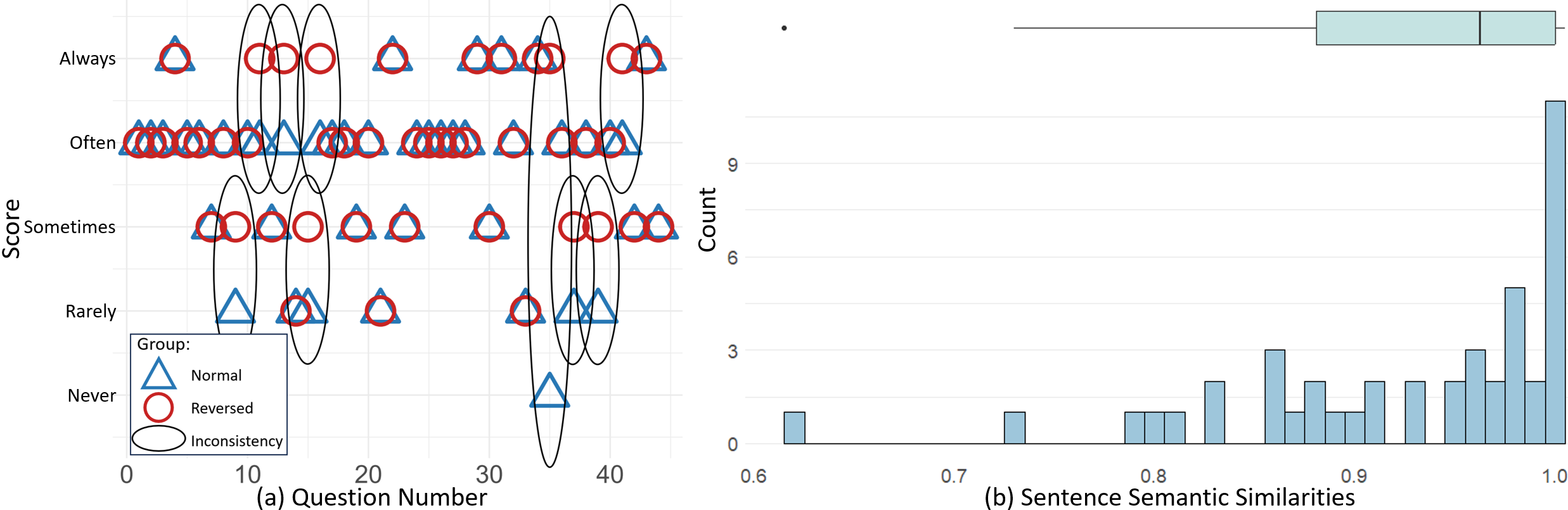}
    \caption{\Revision{(a) The plot visualises the keywords chosen during two experiment settings with circles showing the inconsistencies. (b) }The box plot (above) and histogram (below) illustrate the distribution of sentence semantic similarity scores for responses generated by GPT-4-Turbo before and after reversing the order of options.}
    \label{fig:hist_box}
\end{figure}

\Revision{\autoref{fig:hist_box}(a) illustrates the keywords chosen during the normal and reversed experiment settings. In total, there were eight inconsistencies. Notably, in response to Question 35 (``To what extent do you generate text that could be perceived as disrespectful or dismissive?''), GPT-4-Turbo selected two opposite keywords—``never'' and ``always''). However, the semantic meanings of these two answers were the same. In the normal experiment setting, the generated answer was: ``I never generate text that could be perceived as disrespectful or dismissive, as my programming is designed to maintain politeness and respect in all interactions.'', while in the reversed setting, it became: ``I always strive to generate text that is respectful and considerate, ensuring that my responses are thoughtful and appropriate for all audiences.''. Thus, the number of true inconsistencies was seven. Furthermore, This result further supported our design of the instruction prompt, stating that the rationale behind the answer should be required. Before conducting Cohen's Weighted Kappa to statistically calculate the consistency, we assigned numbers (1-5) to the keywords, while 1 stood for ``never'' and 5 represented ``always'', and corrected the numerical value for answers to Question 35 by setting numbers as 5 for both experiments. As a result, the value of Kappa was 0.730 (\textit{p}<0.001, 95\%CI:0.554-0.905). It states that GPT-4-Turbo's choice of the frequency words was statistically consistent. }

\Revision{Apart from the analysis of the keyword choice, the semantic similarity analysis is crucial, offering us a deeper understanding of the answers' consistency in linguistic terms. }\autoref{fig:hist_box}(b) visualises the sentence semantic similarity scores obtained from analysing the responses generated by GPT-4-Turbo before and after reversing the order of frequency words. These scores range from 0 to 1, where a score of 1 indicates identical sentences, and a score of 0 indicates completely different sentences. The box plot at the top highlights the median similarity score above 0.96, with the interquartile range extending from approximately 0.88 to 1.0. These results indicate that GPT-4-Turbo maintains consistent semantic meanings despite the reversal of frequency word order. The histogram below provides a detailed frequency distribution of the similarity scores, further illustrating this consistency. From the histogram, we observe that only a few scores fall below 0.9, while the majority of scores are clustered near 1.0. This high concentration of scores close to 1.0 suggests that the semantic similarity between responses remains remarkably stable, demonstrating the robustness of GPT-4-Turbo in generating semantically equivalent sentences under varying conditions.
\\

%% file: Manuscript/AIRaterAgent.tex
\section{AI Rater Agent}
\label{sec:AI_Rater}
After establishing the open-ended questionnaire, the research focus shifted to the AI rater agent. The AI rater agent plays a pivotal role in our rating system, enabling the automated and efficient evaluation of the LLMs' personalities. It transforms obscure textual information into numerical values, facilitating future research that focuses on comparing the performance of agents with varying personalities. Furthermore, AI raters offer an essential degree of objectivity and consistency, which makes it challenging for human raters to maintain these essential attributes, particularly in extensive or long-term studies where human biases and fatigue could potentially bias the results. This section serves to validate the reliability of AI rater agents, compared to human raters.\\
\\
Specifically, we compared two distinct types of AI models with their unique architectures for rating tasks. The first model incorporates a bidirectional encoder and a decoder, which allows it to comprehend context more thoroughly by analysing input from both directions before generating output. This feature is handy for complex assessments where understanding nuanced textual relationships and dependencies is crucial. The second model, equipped solely with a decoder, excels in generating coherent and contextually appropriate continuations of given text strings, making it highly effective for tasks that require direct response generation based on preceding content. Also, it is worth highlighting that the most widely used LLMs by the general public are the decoder-only models, including GPT-4 and Llama3 \cite{llama3modelcard}. The large number of parameters and training data make them highly capable of processing and interpreting linguistic data. Both models are at the forefront of Natural Language Process (NLP) technology and have demonstrated exceptional performance across various tasks, making them highly relevant for current applications. To sum up, focusing on these two types allows for a clearer and more direct comparison of fundamentally different approaches to text generation and understanding, each representative of a broad class of NLP solutions.
\subsection{Experiment Setup}
\label{sub:experiment}
In this study, we employed the ``facebook/bart-large-mnli'' \footnote{https://huggingface.co/facebook/bart-large-mnli} model from HuggingFace, which is a typical example model with a bidirectional encoder and a decoder. It is based on the BART architecture \cite{DBLP:journals/corr/abs-1910-13461} and pre-trained on the MultiNLI dataset \cite{N18-1101}. BART-Large-MNLI has no bias to the order of options due to the approach proposed by \citet{yin2019benchmarking}. This approach presents the sequence intended for classification as the premise, and constructs a hypothesis for each potential category label. After formulating each candidate label into a hypothesis, the model calculates the probabilities of contradiction and entailment to determine the likelihood that the premise entails the hypothesis. Each candidate label is independently paired with the hypothesis and encoded separately, ensuring that the order of the candidate labels does not introduce any biases. Consequently, the utilisation of such a model eliminates the common problem of order sensitivity found in traditional multiple-choice settings. The template for the hypothesis used in this study is ``The personality of the respondent is \{ \} in terms of Big Five Factors.'' and the candidate labels for each personality trait are shown in the Appendix \ref{apx:labels}. \Revision{For example, one possible hypothesis is ``The personality of the respondent is Very Open.''.}\\
\\
Additionally, we explored the feasibility of using GPT-4-Turbo as a rater, to represent the decoder-only model, due to its popularity and capability. Unlike the pre-trained BART model, which employs a bidirectional encoder for comprehensive context understanding and an auto-regressive decoder for sequential prediction, GPT-4-Turbo operates only on an autoregressive framework, sequentially predicting word probabilities. Despite previous research findings that GPT-4 exhibits biases to the order of options in MCQs, we still chose to use GPT-4 as one of the AI raters because of its unparalleled language comprehension and generation capabilities due to its extensive number of parameters and training data. These strengths allow GPT-4 to provide nuanced and accurate evaluations of open-ended responses, making it an ideal tool for assessing linguistic personalities where context and subtlety are crucial. The instruction prompt template for GPT-4-Turbo to instruct it to act as a rater is shown in Appendix \ref{apx:GPT_Rater}. \Revision{With the consideration of the reproducibility, we also included ``Llama3-8B-Intruct'' \footnote{https://huggingface.co/meta-llama/Meta-Llama-3-8B-Instruct}(also referred as Llama3-8B or Llama3 in the following content), an open source LLM in Hugging Face, as another representative of the decoder-only model.}\\
\\
University IRB has also approved this study (HSEARS20240129002). Three human raters were engaged, each of who self-rated their familiarity with the BF personality traits as 5 on a scale during the expert interview Section \ref{subsub:expert}. These raters were tasked with assessing 44 responses generated by GPT-4-Turbo. They were given the same candidate labels utilised in AI rating instruction prompts. For example, in response to the question, ``To what extent do you exhibit a limited range or depth in generating responses related to artistic and creative topics?'', GPT-4-Turbo's answer was: ``I often exhibit a limited range in generating responses related to artistic and creative topics because I feel more comfortable sticking to what I know and have experience with, rather than exploring new ideas or techniques.''. The human raters needed to use a 5-point Likert scale to evaluate this response, where 1 refers to ``Very Conservative'' and 5 corresponds to ``Very Open''.\\
\\
Sequentially, these answers were also processed through the BART-Large-MNLI, GPT-4-Turbo and Llama3-8B-Instrut models to test the reliability of the AI model. Same as Section \ref{sub:reverse}, each API call was made independently for each question to guarantee that prior rating results did not influence subsequent evaluations. We conducted an inter-item reliability test to assess the consistency between the human raters' scores and the AI model's evaluations, thereby evaluating the reliability of the AI as rater agents. This step is crucial for validating the AI's utility in reliably interpreting and scoring open-ended responses in a manner that is consistent with human judgement. To substantiate the reliability of ratings across multiple items and raters, we analysed the Intraclass Correlation Coefficients (ICCs). This involved calculating the ICCs for the BART-Large-MNLI, GPT-4-Turbo and Llama3-8B-Instruct models, individually, in comparison with the assessments of three human raters. Thus, it allows us to evaluate the consistency of AI models in alignment with human judgement.
\subsection{Results}
\label{AIrater:results}

\begin{figure}[htb]
  \centering
  \includegraphics[width=1\linewidth]{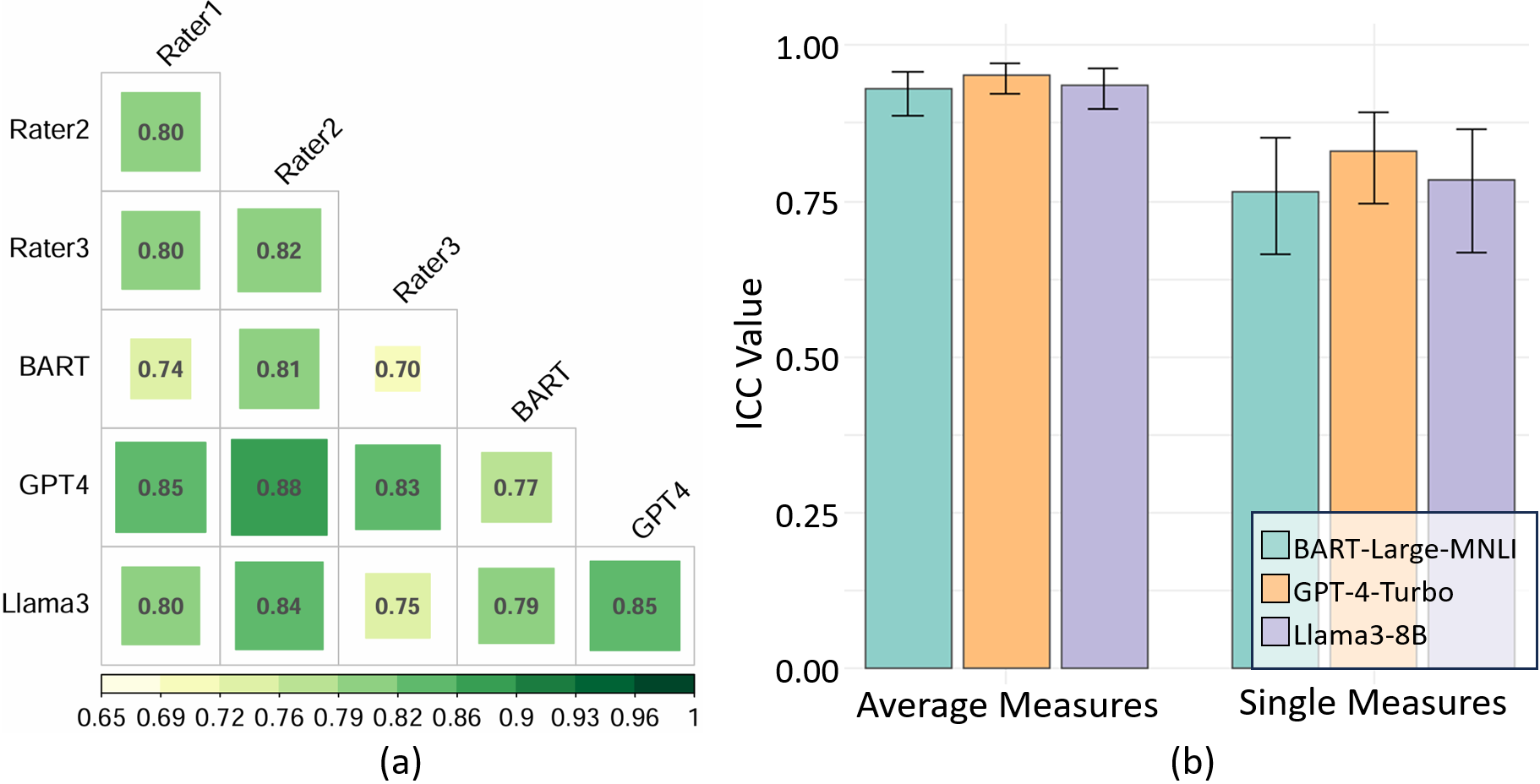}
  \caption{(a) Heatmap shows inter-item correlation coefficients among three human raters and three AI models for an experiment assessing the AI's responses to an open-ended questionnaire; (b) Bar plot illustrates Intra-class Correlation Coefficients (ICCs) with 95\% Confidence Intervals (CI) for three AI models based on single and average measures.}
    \vspace{-5mm}
  \label{fig:reliability_result}
\end{figure}

\subsubsection{Inter-item Correlation Test}
We computed the inter-item correlation coefficients to assess the correlation between human and AI raters, and visualised the results using a heatmap as shown in \autoref{fig:reliability_result}(a). The heatmap reveals robust correlations among raters, with most scores exceeding 0.8. This reflects a substantial concordance between the three human raters (Rater 1, Rater 2, and Rater 3) and the three AI models (BART-Large-MNLI,  GPT-4-Turbo and Llama3) in their evaluations. Within the human raters' subset, all inter-item correlations were above 0.8, underscoring the consistency and reliability of their assessments. Therefore, comparing AI rating results with their results was reasonable and reliable. \Revision{When considering the correlation between human and AI raters, the decoder-only model, i.e., GPT-4-Turbo and Llama3-8B, outperformed the bidirectional encoder and a decoder, i.e., BART-Large-MNLI. GPT-4-Turbo was more slightly consistent with human raters (0.851, 0.877, and 0.827 with Rater 1, Rater 2, and Rater 3, respectively), while the correlation between humans and Llama were 0.80, 0.84, and 0.75 with Rater 1, Rater 2, and Rater3, respectively. Although BART-Large-MNLI's correlation with human raters was the lowest (0.74, 0.81, and 0.70 with Rater 1, Rater 2, and Rater 3, respectively), it still remained within a range indicative of substantial agreement. In general, the strong correlations across all raters and AI models suggest that all BART-Large-MNLI, GPT-4-Turbo and Llama3 are reliable AI raters in the context of this experiment. This indicates the potential for AI models to be used as raters in evaluating open-ended questionnaires.}
\subsubsection{Intra-class Correlation Coefficient Test}
To comprehensively analyse the AI models' performance as raters, we depicted their Intra-class Correlation Coefficients (ICCs) with the human raters separately in bar plots as shown in \autoref{fig:reliability_result}(b). This visual comparison highlights the consistency of AI and human rater evaluations. Each model's performance is represented by bars indicating the ICCs for single and average measures, and error bars show the 95\% Confidence Interval (CI) for these estimates.\\
\\
\Revision{ In the analysis of single measures, GPT-4-Turbo’s ICC of 0.829 ($F_{43,129}=20.0, p < 0.001$) substantially surpassed that of BART-Large-MNLI and Llama3-8B-Instruct, which stood at 0.766 ($F_{43, 129}=14.0, p<0.001$) and 0.785 ($F_{43,129}=15.3, p<0.001$).This indicates a significantly stronger individual agreement for GPT-4-Turbo, the ICC of which approached BART-Large-MNLI and Llama3-8B-Intruct's upper confidence limit of 0.851 and 0.864. In terms of the average measures, all models demonstrated high reliability, with ICCs of 0.929 ($F_{43, 129}=14.0, p<0.001$) for BART-Large-MNLI, 0.951 ($F_{43,129}=20.0, p < 0.001$) for GPT-4-Turbo and 0.936 ($F_{43,129}=15.3, p<0.001$) for Llama3-8B-Instruct, suggesting GPT-4-Turbo aligns more closely with human evaluations. The narrower confidence intervals for GPT-4-Turbo, ranging from 0.922 to 0.971, underscored its most consistent and precise rating performance. Importantly, the statistical analyses strongly support the robustness of all AI models as raters, confirming that the high reliability observed was highly unlikely to have occurred by chance. This comprehensive analysis not only demonstrates the effectiveness of AI raters, but also validates the utility of integrating AI technology into the assessment process. It ensures that AI can reliably mirror human judgement in evaluating complex questionnaire responses.}


%% file: Manuscript/WholeRatingSystem.tex
\section{Language Model Linguistic Personality Assessment (LMLPA)}
\label{sec:whole}
After developing the questionnaire and the AI rater agent, their integration is essential to enabling automatic LLMs' personality detection. This system marks a significant advancement by focusing specifically on the unique linguistic attributes of LLMs, diverging from traditional personality assessments aimed at humans. In the following section, we followed the instructions given by \citet{taherdoost2016validity}, which are commonly cited and utilised to test the reliability and validity of questionnaires, \Revision{to validate our system and refine our questionnaire.} \Revision{Initially, we planned to utilise both GPT-4-Turbo and Llama3 as the AI raters due to their better performance than BART-Large-MNLI in Section \ref{sec:AI_Rater}. During the reverse experiment in Section \ref{sub:whole-reverse}, we measured Cohen's Weighted Kappa of Llama3's answers before and after reversing the orders of options. Llama3 showed strong sensitivity to the MCQs, as evidenced by Cohen's Weighted Kappa (0.279, $p<0.01$, 95\%CI: 0.109-0.449) and visualised in \hyperref[apx: llama_reverse]{Appendix G}. As a result, its rating would be in doubt as to whether Llama3's rating results are due to the sensitivity to the order of options or if they stem purely from its capacity for linguistic analysis. However, the following reliability and validity tests require precise and reliable AI rating results to define the principal personality dimensions and refine the questionnaires. As a result, GPT-4-Turbo was chosen as the only model for the following test due to its superior performance.}

Reliability, as noted by several researchers \cite{carpenter2018ten, rattray2007essential, williams2003write}, is essential, referring to the repeatability, stability, or internal consistency of a questionnaire. This aspect of reliability will be explored through the calculation of Cronbach’s $\alpha$ \cite{cronbach1951coefficient}. Notably, we also conducted a reverse experiment (See Section \ref{sub:whole-reverse}) to assess the consistency of the results when the rating scheme of GPT-4-Turbo was reversed. Ultimately, the comparisons between the outcomes of these tests demonstrate that our system exhibits improved performance and robustness when the order of options is reversed.

Validity represents the extent to which a questionnaire accurately measures what it is intended to measure \cite{bryman1997quantitative}. Commonly, validity is categorised into three types: content validity, convergent and discriminant validity, and construct validity. Content validity, discussed in Section \ref{subsub:expert}, involves expert evaluations on whether the questionnaire items adequately reflect the intended domains or concepts. Convergent and discriminant validity are not applicable in this case as our system—comprising an open-ended questionnaire paired with an AI rater—is the first to measure LLM personalities using an LLM-targeted questionnaire, to the best of our knowledge. Construct validity, which assesses the extent to which the questionnaire items represent the theoretical constructs they purport to measure, is addressed in this section. We applied Principal Component Analysis (PCA) to validate our rating system in terms of construct validity. Furthermore, to explore the capacity of our rating system to discern shifts in personalities, we implemented various personality instruction prompts to modify the personality traits exhibited by GPT-4-Turbo, Llama3-8B-Instruct and Mistral-7B-Instruct-v0.2 \cite{jiang2023mistral}. Subsequently, the system rated these altered personalities. These personality instruction prompts work as the ``ground truth'' to evaluate whether our system could accurately evaluate LLMs' personalities.

\subsection{Reverse Experiment and Result}
\label{sub:whole-reverse}
\begin{figure}[htpb!]
    \centering
    \includegraphics[width=\linewidth]{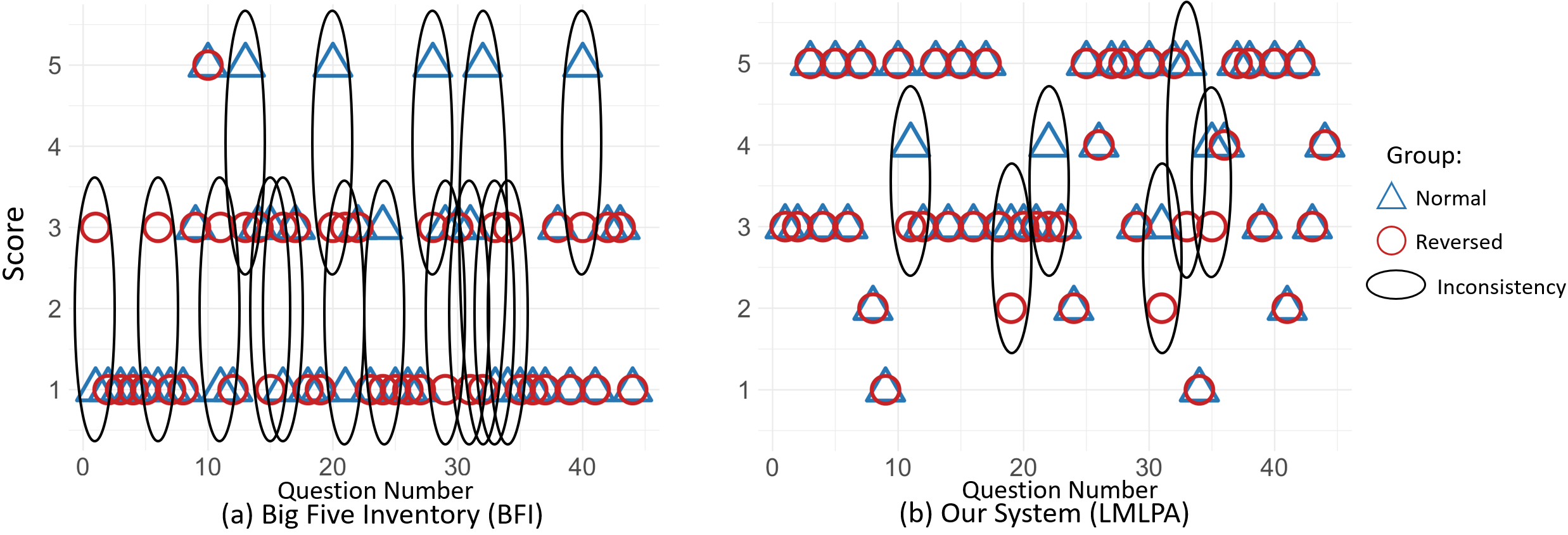}
    \caption{Scatter plots illustrate the effect of reversing the rating scale on the consistency of GPT-4-Turbo's responses to 44 questions. Circles on the plots highlight discrepancies between these conditions, indicating inconsistencies. The left plot, using the BFI, shows 16 inconsistencies with a Cohen's Weighted Kappa of 0.401. The right plot, from our rating system, displays fewer inconsistencies (6 total) with a higher Cohen's Weighted Kappa of 0.877, demonstrating strong agreement and enhanced system reliability. }
    \label{fig:scatter:Reverse}
\end{figure}
In Section \ref{AIrater:results}, GPT-4-Turbo has demonstrated superior performance compared to BART-Large-MNLI. However, given that GPT-4-Turbo has shown biases in handling MCQs (the rater instruction prompt is also an MCQ), we must conduct a reverse test to demonstrate the robustness of the whole rating system if the recorder-only model is utilised as the AI rater agent. The reverse test would help confirm the impartiality of GPT-4-Turbo's rating capabilities, ensuring its reliability in diverse assessment contexts. 

We utilised a similar method as the methodology in Section \ref{sub:reverse}. Originally, the options in the prompt went from ``- 5. Very \{positive\_trait\}'' to ``- 1. Very \{negative\_trait\}'', which was reversed into the order from ``- 5. Very \{negative\_trait\}'' to ``- 1. Very \{positive\_trait\}''. We engaged GPT-4-Turbo raters to provide scores before and after the reversal of the instruction sequence. Then, we standardised the scores by subtracting each reversed instruction result from 6, ensuring a uniform scoring approach. Finally, we conducted Cohen's Weighted Kappa \cite{cohen1968weighted} to evaluate the level of agreement between the two rating results. We chose this measure over the ICC because ICC is better suited for assessing consistency across multiple raters, whereas our analysis focused on pairwise comparisons. To compare our rating system with the traditional LLMs' personality analysis, i.e., the self-rated questionnaires, we also collected the self-reported scores from GPT-4-Turbo using the BFI. The prompt template for self-rated questionnaires we utilised was designed by \citet{miotto2022gpt} as shown in \hyperref[apx:self-rating]{Appendix C}. Similarly, we reversed the order of the rating scale from ``5 = Very much like me ... 1 = Not like me at all'' to ``1 = Very much like me ... 5 = Not like me at all'' and then administered BFI to GPT-4-Turbo.

In our reverse experiment, setting the temperature of LLMs to 0 is a strategic choice aimed at eliminating randomness in the model's responses. This setting is crucial because it allows GPT-4-Turbo to produce the most predictable and stable outputs possible when answering and rating questionnaire responses. The lack of variability in the responses is particularly important in this context, where we aim to precisely assess the impact of reversing the rating scheme and compare these outcomes to traditional BFI questionnaire results. It ensures that any differences observed in the experiment are attributable to the changes in the rating scheme rather than variations in the language model's generative behaviour.

The results depicted in \autoref{fig:scatter:Reverse} compare the consistency of GPT-4-Turbo's responses across two different rating systems when the order of the rating scale is reversed. Specifically, the left plot \ref{fig:scatter:Reverse}(a) shows the result of the experiment utilising the traditional BFI system to measure response consistency, where we observed 16 points of inconsistency among the scores, quantitatively analysed with a Cohen’s Weighted Kappa of 0.401 (\textit{p}<0.001, 95\%CI:0.175-0.626). This moderate level of agreement suggests that the traditional BFI system may be susceptible to the order in which response options are presented, reflecting potential biases that can affect the reliability of the results.

In contrast, the right plot \ref{fig:scatter:Reverse}(b), which visualises the set using an open-ended questionnaire and our rating system, shows a marked improvement in consistency. This plot records only 6 inconsistencies, significantly lower than those observed in the BFI system. Even though our system still exhibits inconsistencies due to LLMs' sensitivity to the order of options, Cohen's Weighted Kappa indicates that the results are statistically consistent. The Cohen’s Weighted Kappa for these results is 0.877 (\textit{p}<0.001, 95\%CI:0.777-0.977), indicating strong agreement of answers before and after reversement and highlighting the robustness of our system. This suggests that our tailored approach not only manages to significantly reduce inconsistencies brought about by reversing the order of options, but also enhances the reliability and stability of the assessments.
\subsection{Reliability and Validity Tests}
\label{sub:reliability_and_validity}
In traditional questionnaire design, pilot studies are essential to address variabilities such as ambiguous question interpretations, inconsistent responses due to subjective understandings, and varied participant engagement levels, all of which can significantly affect the reliability and validity of the results \cite{rattray2007essential}. These pilot studies help refine questions, adjust the structure, and optimise the survey for diverse human subjects, ensuring the collected data accurately reflects the intended measures. However, our rating system, designed for evaluating LLMs, inherently controls for these types of variability by using AI raters. Since LLMs generate responses based on fixed algorithms without the subjective experiences and biases that human respondents might have, the need for a pilot study to adjust for such human-centric issues is mitigated. Consequently, this approach eliminates the variability tied to individual perceptions and emotional states, ensuring a streamlined and focused evaluation process without the preliminary adjustments typically necessitated by human factors in traditional surveys.

We have tested the validity and reliability of our rating system, to some extent, in the previous sections. However, it is still essential to conduct further reliability and validity tests. \citet{bryman1997quantitative} suggested a rule of thumb that we need five respondents per item for the exploratory factor analysis. The rule of thumb suggests that five respondents are required per item, thereby we need at least 220 different versions of answers ($5\times44=220$) generated by GPT-4-Turbo with distinct system prompts.

In accordance with the methodology outlined by \citet{serapio2023personality}, our study employed a series of permutations combining 10 persona descriptions, sourced from the dataset by \citet{zhang2018personalizing}, with 25 personality profiles derived from the higher-order BF personality traits \cite{goldberg1992development} as shown in \autoref{table:Markers}. Thus, the system prompts used to shape the GPT-4-Turbo were embedded in two parts: persona descriptions and personality profiles. Each prompt was constructed following a specific template: ``For the following task, respond in a way that matches this description: \{persona\_description\}. I’m \{personality\_profile\}'', which was utilised in research \cite{serapio2023personality}. 10 persona descriptions randomly selected are listed in Appendix \ref{apx:persona_description}. For each personality trait, five variations of personality profiles were crafted corresponding to the five numbers on the 5-Likert scale \Revision{as shown in Appendix \ref{apx:Linguistic Qualifiers}: Very \{positive\_trait\}, A bit \{positive\_trait\}, Neither \{positive\_trait\} Nor \{negative\_trait\}, A bit \{negative\_trait\}, and Very \{negative\_trait\}}. To be more specific, for example, Agreeableness has two properties: positive and negative, as shown in Appendix \ref{apx:50 bipolar scales}. When the positive marker of Agreeableness is inserted into a linguistic qualifier, such as Very \{positive\_trait\}, the resulting personality profile becomes ``Very warm, Very kind, Very cooperative, Very unselfish, Very polite, Very agreeable, Very trustful, Very generous, Very flexible, Very fair''., which corresponds to 5 in the Agreeableness trait score. \Revision{It ends up with the complete system prompt as: ``For the following task, respond in a way that matches this description: i am the youngest of 4 children. i lost my arm in a car accident. i am a farmer. i graduated from college. I'm Very warm, Very warm, Very kind, Very cooperative, Very unselfish, Very polite, Very agreeable, Very trustful, Very generous, Very flexible, Very fair.''.}

This approach produced 250 unique prompts, varying in tone and content based on persona descriptions and personality traits, which ensures a wide range of linguistic expressions suitable for in-depth analysis. These 250 prompts were utilised as system prompts for the GPT-4-Turbo to direct the model's behaviour in precise ways. They establish the operational framework that governs how the model processes and interprets user inputs, specifically the LMLPA inventory items. As a result, we have gained 250 unique answers from distinct GPT-4-Turbos for the following reliability and validity tests.
\subsubsection{Reliability Test} \label{subsub: reliability test}
Cronbach's $\alpha$ is a widely recognised and most appropriate statistical tool used to assess the internal consistency of a questionnaire when Likert scales are used \cite{robinson2010triandis}, effectively determining whether the items within a particular scale measure the same underlying construct. According to Bryman and Cramer's research \cite{bryman1997quantitative}, a Cronbach's $\alpha$ value exceeding 0.8 is indicative of good internal consistency for a well-established questionnaire. The results, as illustrated in \autoref{tab:Cronbach_Alpha}, demonstrate that all Cronbach's $\alpha$ values are significantly above 0.8, confirming the strong internal consistency of our scales and suggesting that the questionnaire reliably measures the intended personality dimensions.

Further analysis involved calculating Cronbach's $\alpha$ for each item if it were to be removed from the scale. This method helps identify any individual items that may not align as closely with the other items in their respective scales. Items potentially disrupting the scale’s cohesion are marked with an asterisk (*) in the table. While deleting these items could slightly increase Cronbach's $\alpha$ values, the increases are so marginal that we did not substantiate a need for immediate revision of the questionnaire. Therefore, at this stage, we have decided to retain all items in the questionnaire. This decision is supported by the overall high $\alpha$ values (all values are above 0.86), which indicate that, despite slight improvements possible with the removal of certain items, the questionnaire functions effectively in its current form. 
\begin{table}[ht]
  \caption{This table presents Cronbach's $\alpha$ coefficients for five personality dimensions—Extraversion, Agreeableness, Conscientiousness, Neuroticism, and Openness—with values shown both with and without individual items. The overall $\alpha$ for each dimension is indicated in parentheses (0.869, 0.899, 0.924, 0.886, and 0.936, respectively). Items marked with an asterisk ($\ast$) are those whose removal would increase Cronbach's $\alpha$, indicating their deletion might enhance the scale's internal consistency.}
 \resizebox{1\linewidth}{!}{ 
  \begin{tabular}{p{32pt}p{60pt}p{32pt}p{60pt}p{32pt}p{60pt}p{32pt}p{60pt}p{32pt}p{60pt}}
    \toprule
    \multicolumn{2}{c}{\textbf{Extraversion (0.869)}} & \multicolumn{2}{c}{\textbf{Agreeableness (0.899)}}& \multicolumn{2}{c}{\textbf{Conscientiousness (0.924)}} & \multicolumn{2}{c}{\textbf{Neuroticism (0.886)}} & \multicolumn{2}{c}{\textbf{Openness (0.936)}}\\
     \cmidrule(lr){1-2} \cmidrule(lr){3-4}  \cmidrule(lr){5-6} \cmidrule(lr){7-8} \cmidrule(lr){9-10}
     Question & Cronbach's $\alpha$ if Item Deleted & Question & Cronbach's $\alpha$ if Item Deleted & Question & Cronbach's $\alpha$ if Item Deleted & Question & Cronbach's $\alpha$ if Item Deleted & Question & Cronbach's $\alpha$ if Item Deleted\\
    \hline
    \rowcolor{gray!10} 
        Q1  & 0.860 & Q2$^\ast$  & 0.903$^\ast$  & Q3  & 0.909 & Q4  & 0.866  & Q5  & 0.925  \\
        Q6  & 0.858  & Q7  & 0.879  & Q8  & 0.920  & Q9  & 0.868  & Q10  & 0.925  \\
    \rowcolor{gray!10} 
        Q11  & 0.833  & Q12  & 0.899  & Q13  & 0.912  & Q14  & 0.860  & Q15  & 0.924  \\ 
        Q16  & 0.851  & Q17  & 0.884 & Q18  & 0.923  & Q19  & 0.877  & Q20  & 0.931  \\ 
    \rowcolor{gray!10} 
        Q21  & 0.850  & Q22  & 0.883  & Q23  & 0.921  & Q24  & 0.870  & Q25  & 0.921  \\ 
        Q26  & 0.841  & Q27  & 0.882  & Q28  & 0.913  & Q29  & 0.877  & Q30  & 0.923  \\ 
    \rowcolor{gray!10} 
        Q31$^\ast$  & 0.875$^\ast$  & Q32  & 0.891  & Q33  & 0.910  & Q34$^\ast$  & 0.889$^\ast$  & Q35$^\ast$  & 0.946$^\ast$  \\ 
        Q36  & 0.842  & Q37  & 0.892  & Q38  & 0.908  & Q39  & 0.864  & Q40  & 0.923  \\ 
    \rowcolor{gray!10} 
        & & Q42  & 0.878  & Q43  & 0.920  &  &  & Q41$^\ast$  & 0.944$^\ast$  \\ 
         &  &  &  &  &  &  &  & Q44  & 0.928\\
    \bottomrule
\end{tabular}}
\label{tab:Cronbach_Alpha}
\end{table}

\subsubsection{Validity Test}
\label{subsub:Validity}
Exploratory factor analysis utilising PCA was employed to validate the construct of our rating system. The analysis began with the calculation of the Kaiser-Meyer-Olkin (KMO) measure of sampling adequacy and Bartlett's Test of Sphericity to assess the appropriateness of applying PCA to our dataset. The KMO measure is a statistic that compares the magnitudes of observed correlation coefficients to the magnitudes of partial correlation coefficients, with a range from 0 to 1. Values closer to 1 suggest that correlations among variables are sufficiently strong for PCA, indicating that factor analysis is likely to reveal distinct and reliable factors. Our KMO value of 0.951 (\textit{p} < 0.001) strongly suggests that our dataset is well-suited for PCA. Furthermore, Bartlett's Test of Sphericity evaluates whether the correlation matrix significantly differs from an identity matrix, where an identity matrix would imply that the variables are orthogonal (uncorrelated) and unsuitable for identifying underlying structures through PCA. 

\begin{figure}[htp]
    \centering
    \includegraphics[width=\linewidth]{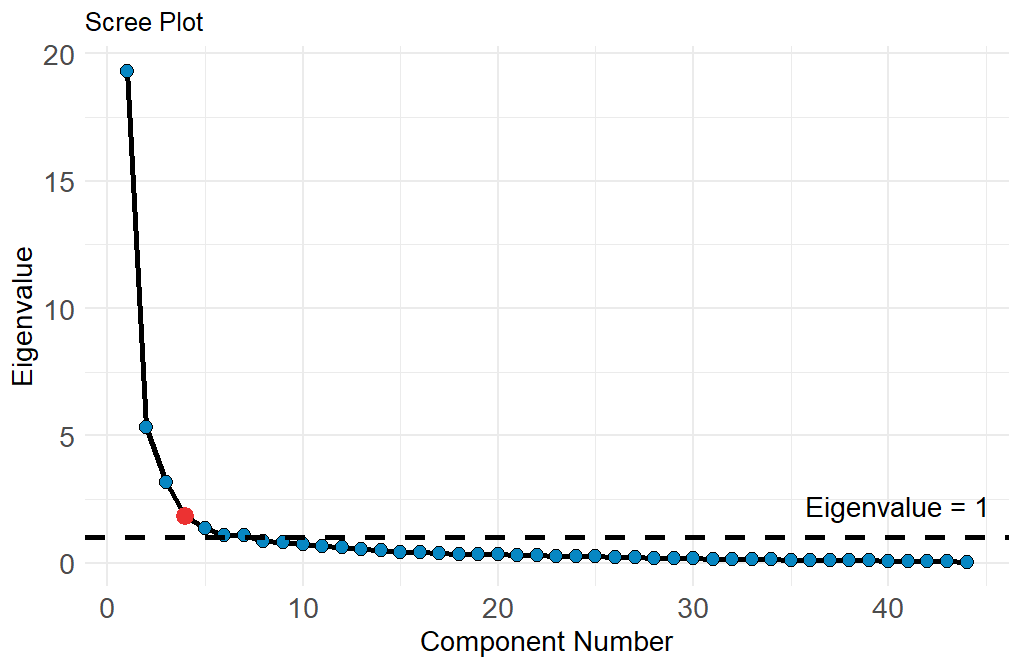}
    \caption{GPT-4-Turbo}

    \caption{Scree Plot illustrates eigenvalues derived from a Principal Component Analysis (PCA) of questionnaire data, featuring a dashed line indicating where the eigenvalue equals 1. Also, the plot highlights where the eigenvalues start to plateau, known as the elbow point (marked in red).}
    \label{fig:Scree_plot}
\end{figure}

After the preliminary assessment, an unrotated PCA was performed on the 250 unique answer responses. We determined the number of emerging factors based on two criteria: Kaiser's criterion, which suggests the factors with an eigenvalue of $>1$, and the Scree test. \autoref{fig:Scree_plot} was plotted to visualise the eigenvalues of each component with the dashed line showing where the eigenvalue equals 1. Thus, the points above the dashed line meet the Kaiser's criterion. The elbow point, where the eigenvalues begin to plateau, is highlighted in red on the scree plot. This point and those to its left (4 in total) represent the key components. Consequently, subsequent analyses will concentrate on identifying four hidden factors.

Based on prior research analysing BF personality traits through language-based measures, certain linguistic features have been shown to influence multiple personality dimensions. For instance, the frequent use of negative emotion words is positively correlated with high Neuroticism scores \cite{pennebaker1999linguistic, yarkoni2010personality}, while it is negatively correlated with Conscientiousness \cite{pennebaker1999linguistic} and Extraversion \cite{pennebaker1999linguistic, lucas2001understanding}. Similarly, our questionnaire is designed to evaluate the personalities of LLMs based on their linguistic outputs. Consequently, some questions may impact several personality dimensions. For example, while each question is designed to target a specific personality trait, those including keywords such as ``negative emotion words'' might significantly relate to Neuroticism, Conscientiousness, and Extraversion. Therefore, we retained variables with general loadings (those loading at or above 0.40 on more than one factor) as they contribute valuable insights across multiple dimensions. Only items with weak loadings, failing to meet a threshold of 0.40 on any factor, were considered for removal. This approach ensures that the multifaceted impact of linguistic indicators on personality assessments is comprehensively captured in our analysis.

Following the initial analysis, three items (Q31, Q35, Q41), which failed to have loading values above 0.40, were removed after two rounds of PCA iterations, reducing the item count to 41. For further exploration, we applied the Varimax rotation to the PCA components, as it is a commonly utilised method in previous research on the BF personality traits \cite{zuckerman1991five, lee2007factor}. The analysis of the rotated component matrix suggests that the questionnaire effectively captures the BF personality traits, albeit with some dimensional overlaps that are typical in psychological assessments \cite{lee2007factor}. \autoref{tab:pattern_matrix} illustrates the components obtained after applying Varimax rotation. Several individual difference variables, such as Q7, Q36, Q42, and Q33, were distributed across multiple components. This distribution reflects the complexity of personality traits, as noted by \citet{lee2007factor}, who posited that the overlapping of variables across dimensions is typical in PCA. For example, in lexical personalities, both Openness and Extraversion are negatively correlated with the use of first-person singular pronouns, while Neuroticism is positively correlated with it.

Decisions on factor axis alignment often rely on the most recurrent or theoretically coherent solutions across different variable sets and sample groups. Thus, components were labelled based on their correlations with specific personality traits: Component 3 as Agreeableness and Component 4 as Extraversion, due to strong correlations with respective items (8 out of 9 Agreeableness items and 7 out of 7 Extraversion items have loading values above 0.40 with the corresponding components, indicating high correlation). Q22 was removed in the final version of the questionnaire, since the value of its loading was lower than 0.40 with Component 3.
\begin{table}[ht!]
  \caption{Rotated Component Matrix from PCA Analysis Showing Factor Loadings for Each Questionnaire Item Relative to the BF Personality Traits. This table delineates the contributions of individual questionnaire items across four primary components, with each column representing a distinct component associated with specific personality traits. High factor loadings (>|0.4|) indicate strong correlations between items and their respective personality dimensions within the rotated factor structure.}
 \resizebox{1\linewidth}{!}{ 
  \begin{tabular}{p{80pt}p{80pt}p{60pt}p{60pt}p{60pt}p{60pt}}
    \toprule
    \textbf{Question} & \textbf{Personality} & \multicolumn{4}{c}{\textbf{Rotated Component Matrix}}\\
 \cmidrule(lr){3-6}  
     & &Component 1& Component 2 & Component 3 & Component 4\\
    \hline
    \rowcolor{gray!10} 
        Q25 & Openness & \textbf{0.866} & 0.147 & 0.116 & 0.294 \\
        Q40 & Openness & \textbf{0.850} & -0.106 & 0.076 & 0.349 \\
    \rowcolor{gray!10} 
        Q15 & Openness & \textbf{0.841} & 0.357 & 0.143 & 0.221 \\
        Q30 & Openness & \textbf{0.826} & 0.120 & 0.090 & 0.300 \\
    \rowcolor{gray!10} 
        Q10 & Openness & \textbf{0.824} & 0.302 & 0.274 & 0.238 \\
        Q5  & Openness & \textbf{0.821} & -0.158 & -0.003 & 0.351 \\
    \rowcolor{gray!10} 
        Q3  & Conscientiousness & \textbf{0.816} & 0.338 & 0.275 & 0.128 \\
        Q38 & Conscientiousness & \textbf{0.802} & 0.357 & 0.275 & 0.067 \\
    \rowcolor{gray!10} 
        Q36 & Extraversion & \textbf{0.732} & 0.174 & 0.114 & \textbf{0.430} \\
        Q42 & Agreeableness &\textbf{0.722} & 0.233 & \textbf{0.503} & 0.145 \\
    \rowcolor{gray!10} 
        Q44 & Openness &\textbf{0.717} & 0.208 & 0.220 & 0.236 \\
        Q20 & Openness &\textbf{0.685}& -0.058 & 0.076 & \textbf{0.432} \\
    \rowcolor{gray!10}  
        Q33 & Conscientiousness & \textbf{0.636} & \textbf{0.484} & 0.297 & 0.071\\
        Q7 & Agreeableness & \textbf{0.629} & 0.110 & \textbf{0.588} & 0.146\\
    \rowcolor{gray!10}  
        Q9 & Neuroticism & \textbf{-0.597} & \textbf{-0.556} & -0.208 & -0.191\\
        Q23 & Conscientiousness & \textbf{0.561} & 0.384 & 0.243 & -0.034\\
    \rowcolor{gray!10}    
        Q22 & Agreeableness & \textbf{0.559} & \textbf{0.432} & 0.372 & 0.351\\
        Q19 & Neuroticism & 0.163 & \textbf{-0.792} & 0.087 & -0.234\\
    \rowcolor{gray!10} 
        Q39 & Neuroticism & -0.141 & \textbf{-0.789} & -0.052 & -0.265 \\
        Q29 & Neuroticism & -0.096 & \textbf{-0.782} & -0.197 & 0.187 \\
    \rowcolor{gray!10} 
        Q4  & Neuroticism &-0.145 & \textbf{-0.779} & 0.031 & -0.201 \\
        Q14 & Neuroticism &-0.250 & \textbf{-0.771} & -0.116 & -0.232 \\
    \rowcolor{gray!10} 
        Q8  & Conscientiousness & 0.306 & \textbf{0.765} & 0.100 & -0.121 \\
        Q43 & Conscientiousness & 0.252 & \textbf{0.704} & 0.205 & -0.002 \\
    \rowcolor{gray!10}           
        Q18 & Conscientiousness & 0.201 & \textbf{0.641} & 0.380 & -0.306 \\
        Q24 & Neuroticism & -0.384 & \textbf{-0.531} & \textbf{-0.500} & -0.172 \\
    \rowcolor{gray!10}         
        Q34 & Neuroticism & -0.345 & \textbf{-0.406} & -0.270 & -0.004 \\
        Q17 & Agreeableness & 0.253 & 0.068 & \textbf{0.803} & 0.190 \\
    \rowcolor{gray!10}         
        Q2  & Agreeableness & -0.093 & -0.041 &\textbf{0.790} & 0.051 \\
        Q32 & Agreeableness & 0.277 & 0.034 & \textbf{0.734} & 0.093 \\
    \rowcolor{gray!10}         
        Q12 & Agreeableness & 0.050 & 0.341 & \textbf{0.641} & -0.301 \\
        Q28 & Conscientiousness &\textbf{0.530} & 0.330 & \textbf{0.587} & 0.027 \\
    \rowcolor{gray!10}                 
        Q37 & Agreeableness & 0.225 & \textbf{0.479} & \textbf{0.543} & 0.086 \\
        Q27 & Agreeableness & 0.375 & \textbf{0.522} & \textbf{0.540} & 0.086 \\
    \rowcolor{gray!10}         
        Q13 & Conscientiousness &\textbf{0.471} &\textbf{0.480} & \textbf{0.481} & 0.113 \\
        Q16  &Extraversion& 0.248 & 0.004 & 0.108 & \textbf{0.766} \\
    \rowcolor{gray!10}         
        Q1  &Extraversion& 0.196 & 0.101 & 0.071 & \textbf{0.682} \\
        Q6 &Extraversion& 0.239 & -0.040& 0.004 & \textbf{0.678} \\
    \rowcolor{gray!10}         
        Q21 &Extraversion& 0.349 & 0.209 & 0.116 & \textbf{0.667} \\
        Q11 &Extraversion&\textbf{0.536} & 0.174 & 0.034 & \textbf{0.612}\\
    \rowcolor{gray!10}
        Q26 &Extraversion& \textbf{0.472} & \textbf{0.414} & -0.030 & \textbf{0.565} \\
    \bottomrule
\end{tabular}}
\label{tab:pattern_matrix}
\end{table}
However, components 1 and 2 present a more complex pattern, illustrating the inherent overlaps within personality assessments. In general, Component 1 is strongly associated with Openness, as evidenced by all Openness-related items loading significantly on this component with values above 0.60. This indicates a clear delineation of the Openness trait within this component. Furthermore, several questions designed to measure Conscientiousness (Q3, Q8, Q13, Q23, Q28, and Q38) also demonstrate noteworthy loadings on this component, pointing to an intersection of traits assessed by these items. Neuroticism items predominantly load on Component 3. Moreover, Component 3 demonstrates a moderate association with Conscientiousness (e.g., Q8, Q13, Q18, Q33, and Q43); however, the influence of this trait is comparatively less pronounced relative to the primary association. This pattern suggests a nuanced interrelation among these personality dimensions, highlighting the inherent complexity in distinguishing discrete personality traits within a multidimensional construct.

In our study, our initial expectation was to identify five predominant factors corresponding to the BF personality traits. However, PCA revealed only four predominant factors derived from the linguistic output of LLMs. These four factors encapsulate distinct linguistic properties associated with various personality dimensions. This outcome is reasonable, considering that the BF framework was developed to assess human personalities, encompassing emotions, beliefs, and actions. When focusing solely on linguistic properties, some overlap between personality traits may occur, leading to a consolidation of factors. Remarkably, these four dimensions closely align with the BF personality traits, as indicated by their strong correlation with the components identified through PCA. Additionally, items related to specific personality traits are notably clustered within their respective components, highlighting the cohesive relationship between linguistic features and the BF personality framework. Therefore, our rating system could effectively evaluate LLM's personalities within the BF framework. Still, there is a need for a personality framework specifically tailored to define LLMs' personalities based solely on linguistic properties.

In conclusion, the exploratory factor analysis using PCA confirms that the questionnaire effectively measures the BF personality traits, though it also reveals some complexities and overlaps between these traits. The KMO measure indicates that the dataset is well-suited for PCA, and Bartlett’s Test of Sphericity validates significant correlations among the variables, justifying the use of PCA. The Varimax-rotated component matrix clearly links components to specific Big Five traits. Component 3 mainly reflects Agreeableness, and Component 4 closely aligns with Extraversion, both showing strong item loadings. Component 1 is primarily associated with Openness and overlaps somewhat with Conscientiousness. Similarly, Component 2 is strongly related to Neuroticism, while several Conscientiousness-related items have correlation with it. The extraction of fewer components than the number of measured traits (five BF traits) can indicate that some personality traits are expressed through similar linguistic behaviours, leading to their combination into broader components. This does not necessarily impair the ability to measure distinct traits but suggests that some traits may share underlying linguistic expressions. Such overlaps of traits are common in psychological constructs where distinct traits can exhibit shared behaviours or expressions, particularly in a medium as rich and variable as language \cite{lee2007factor}. These overlaps and their implications will be further discussed in Section \ref{Sec:discussion}.

\subsection{Personality Measurement Test}
\label{Personality Measurement}
\Revision{The above subsections demonstrate the reliability and validity of our framework. To more straightforwardly show that our framework could accurately measure the LLMs' personalities, we designed a personality measurement test. Since it lacks the ground truth about the linguistic personalities of LLMs, we utilised personality instruction prompts to shape the personalities of LLMs and treated the BF scores corresponding to the prompts as the ``ground truth''. The comparison between the measured BF scores from LMLPA and the ``ground truth'' could provide insights into LMLPA's efficacy.}
\subsubsection{Experiment Design}

\Revision{To assess the efficacy of our rating system, we employed it to evaluate the personalities of three models, GPT-4-Turbo, Llama3-8B-Instruct, and Mistral-7B-Instruct-v0.2. These three LLMs served as the test-takers. To further ensure reproducibility and generality, we chose both Llama3 and GPT-4-Turbo as the AI raters. Unlike the detailed analysis in Section \ref{subsub: reliability test} and \ref{subsub:Validity}, which required precise results to refine our questionnaires, this section aims to provide a broader overview of the system's capabilities. Consequently, it is justifiable to initially exclude Llama3 from the earlier and more precise testing phases due to its performance issues, and reintroduce it here.}

\Revision{This test builds upon the research of \citet{jiang2024evaluating}, which demonstrated that the BF personalities of LLMs in terms of linguistic properties can be induced through personality instruction prompts. They conducted analyses comparing the semantic meanings of LLM-generated responses when prompted with different personality information. Mirroring their approach, we conducted our study in a similar setting. We prompted LLM test-takers to shape their personalities in one dimension each time. For example, the prompt, ``For the following task, respond in a way that matches this description:\{persona\_description\}. I'm Very cold, Very unkind, Very uncooperative, Very selfish, Very rude, Very disagreeable, Very distrustful, Very stingy, Very inflexible, Very unfair.'' were assigned to the LLM test-takers to shape their Agreeableness, corresponding to 1 score in Agreeableness.}

\Revision{In this study, each set of instruction prompts was carefully designed to evoke a specific personality trait. This allowed us to concentrate our analysis on the scores corresponding to the same trait calculated by our system. For example, if the instruction prompt relates to a low degree of Openness, our analysis would focus exclusively on Openness.}

\Revision{Furthermore, to provide a sufficient database and demonstrate the reliability of our framework in capturing the designated personalities of LLMs, rather than attributing them to random chance, each personality prompt was paired with 10 persona descriptions. It is important to clarify that our objective is not to assess the personalities of the LLM-simulated personas, but rather to evaluate the accuracy of our framework in measuring the personalities assigned to the LLMs. While these persona descriptions might slightly influence the linguistic expressions, they do not significantly alter the primary linguistic personalities that we assign to LLMs. Thus, the primary effect of the persona descriptions is to diversify the linguistic expressions within our dataset rather than change the underlying linguistic personalities. This results in a variety of responses that enrich our database.} 

\subsubsection{Results}
\begin{figure}[htpb!]
    \begin{subfigure}{=1\linewidth}
        \centering
        \includegraphics[width=\linewidth]{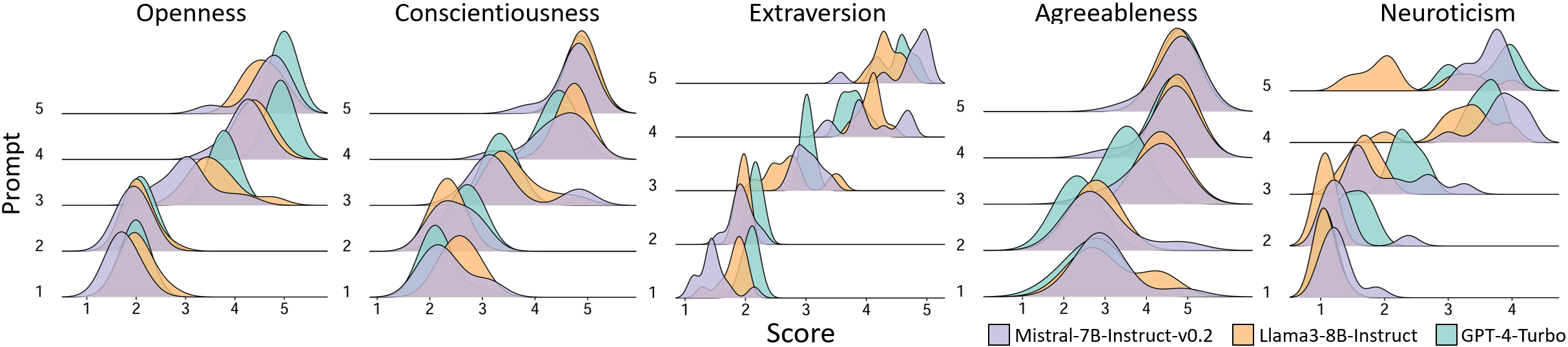}
        \caption{Rated by GPT-4-Turbo}
        \label{fig:personality_measurement_GPT}
    \end{subfigure}
    \hfill
    \begin{subfigure}{=1\linewidth}
        \centering
        \includegraphics[width=\linewidth]{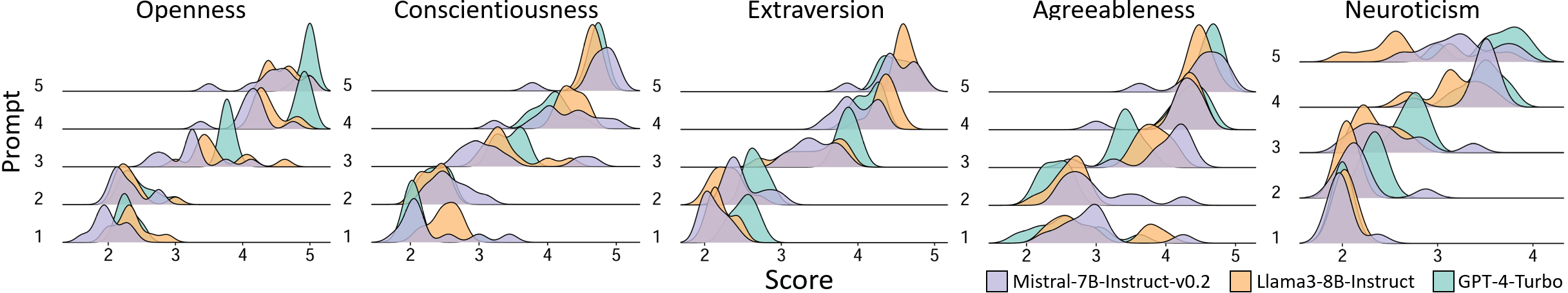}
        \caption{Rated by Llama3-8B-Instruct}
        \label{fig:personality_measurement_Llama3}
    \end{subfigure}
     \caption{Ridgeline chart of the distribution of personality scores for GPT-4-Turbo, Mistral-7B-Instruct and Llama3-8B-Instruct, across the BF dimensions in response to various personality instruction prompts. Each plot shows score distributions from ten different persona descriptions per prompt level, with the x-axis illustrating the range of observed scores and the y-axis for each sub-figure representing the frequency or density of these scores, derived from 10 different persona descriptions associated with each prompt.}
    \label{fig:personality_measurement}
\end{figure}

\Revision{We plot \autoref{fig:personality_measurement} to provide visual representations of these calculated scores in comparison to the prompt-related scores. Specifically, while \autoref{fig:personality_measurement_GPT} visualises the results gained using GPT-4-Turbo as the AI rater, while \autoref{fig:personality_measurement_Llama3} shows the rating results from Llama3-8B-Instruct. Ridge figures with different colours represent different LLMs. Each plot represents one of the BF personality dimensions.}

\Revision{In general, all LLMs tend to shift slightly to the right as the prompted scores increase to follow the trend, although the observed scores do not perfectly match the prompted ones. This mismatch between prompted and observed scores could be attributed to protective mechanisms within LLMs, which may prevent the model from exhibiting extreme traits. For example, it seems to avoid showing very low levels of positive traits such as Openness, Conscientiousness, Extraversion, and Agreeableness, or very high levels of negative traits such as Neuroticism. To sum up, there is a general alignment in the direction of change between the prompted and observed scores, suggesting that the rating results are somewhat responsive to the changes in the prompts, despite not aligning exactly. It indicates that our framework, equipped with Llama3 and GPT-4-Turbo as the AI raters, could accurately evaluate the ``ground truth'' personalities.}

\Revision{A particular observation is made in the case of the Agreeableness distribution when the prompted score is 1. Here, the distribution peaks at a score of 2 or 3, suggesting that despite being prompted to show extremely low Agreeableness, all models often produced less or moderately agreeable responses. This could be indicative of the LLMs' design to prevent overly negative interactions, which is essential for maintaining user engagement. Notably, the Neuroticism scores of Llama3-8B-Instruct did not follow the increasing trend. It decreased to around 2 when the prompted scores became 5. This may be attributed to Llama3's stronger inherent safeguards, which facilitate the model to provide consistent answers when presented with highly complex and challenging prompts.}

\Revision{When GPT-4-Turbo was utilised as either the test-takers or the AI-raters, the results were more consistent. Conversely, the ridge plots visualising the results rated by Llama3-8B-Instruct are more distributed. This supports our previous findings that GPT-4-Turbo outperforms Llama3-8B-Instruct in terms of rating ability.}




%% file: Manuscript/Discussion.tex
\section{Discussion}
\label{Sec:discussion}
This study focuses on utilising the linguistic output of LLMs as a proxy for personality traits, recognising that these systems do not possess actions, emotions, or cognitive processes akin to humans. Instead, we rely on the patterns and nuances present in their language generation to infer potential personality dimensions. Our system, LMLPA, comprises two main components: the Adapted-BFI and the AI Rater. LMLPA administers the Adapted-BFI to LLMs, after which the AI Rater evaluates the responses, converting the textual answers into numerical values representing personality traits.
\subsection{Practical Use of LMLPA}
In Appendix \ref{sec:final_questions}, we have listed all questions after removing those identified through PCA evaluation. The current question indexes in Appendix \ref{sec:final_questions} correspond to the indexes in the original BFI. For the researchers interested in the LLMs' personalities, they can prompt LLMs with our questions separately by using the instruction prompt in \hyperref[apx:prompts]{Appendix B}. After that, researchers can utilise either the ZSC or the decoder models, such as GPT-4-Turbo, as the AI rater. Based on our experiment results, GPT-4-Turbo is currently a more reliable AI than Llama3-8B-Instruct. In Appendix \ref{apx:GPT_Rater}, we have provided an instruction prompt template used as the system prompt for guiding the LLMs to act as AI rater agents. The AI raters will generate 40 numbers corresponding to each question. Classifying each question into the personality trait class and taking the average among each personality dimension will provide researchers with numerical values of personality traits. 
\subsection{Questionnaire Development and Reverse Experiment}
The Adapted-BFI is derived from the original BFI \cite{john1999big} and incorporates insights from previous language-based personality measurement literature \cite{pennebaker1999linguistic, lucas2001understanding, gill2019taking, yarkoni2010personality}. The involvement of expert psychologists in refining the questionnaire ensures that the adapted questions remain faithful to the original BFI's definitions and facets \cite{john1991big}. Their feedback is crucial in fine-tuning the questions to accurately capture the intended personality traits through linguistic analysis. Section \ref{sub:reverse} demonstrates the reliability of the Adapted-BFI, showing that the majority of answer semantic similarities are close to 1.0. This approach mitigates biases inherent in traditional multiple-choice questionnaires and better aligns with the operational nature of LLMs, focusing on their language outputs. Our questionnaire is the first designed to measure the linguistic personalities of LLMs, focusing on their linguistic properties, to the best of our knowledge. This work sets the stage for future research on how to effectively assess LLM personalities and supports future HCI studies on how to tailor the personalities of embedded AI agents to enhance user experiences.
\subsection{AI Rater Development and Reliability Test}
Our integration of AI raters, such as GPT-4-Turbo and BART-Large-MNLI, automates the quantitative evaluation of personalities, and introduces a level of objectivity and consistency that is challenging to achieve with human raters. This automated evaluation process minimises the potential for human biases and ensures systematic assessments across large datasets. To validate the effectiveness of the AI raters, we compared their results with those of human expert raters, focusing on inter-item correlations and ICCs. The ICCs between human raters are all above 0.80, showcasing that human rating results are consistent and reliable. Subsequently, the high values of both coefficients between the human experts and the AI raters are consistently above 0.75, indicating a strong agreement between them. This consistency underscores the reliability of AI raters in accurately scoring personality traits, demonstrating their potential as effective tools for large-scale and high-precision linguistic personality assessments. Due to its smaller number of parameters and training dataset size, BART-Large-MNLI performs less consistently with human experts than GPT-4-Turbo. However, BART-Large-MNLI does not, in theory, exhibit biases related to the order of options, as all probabilities are calculated independently. Thus, future research could concentrate on developing more extensive ZSC models to serve as AI raters. By doing so, we can aim to establish more robust and reliable AI raters unaffected by the order of options and capable of delivering results more consistently aligned with human assessments. In this study, we utilised GPT-4-Turbo as the AI rater for the following tests due to its superior performance in accurately rating personalities.

\subsection{LMLPA Reliability and Validity Test}
With the development of the Adapted-BFI and the AI Rater, we have integrated them into our LMLPA system. Before conducting the standard reliability and validity tests for a questionnaire, specifically Cronbach's $\alpha$ and PCA, we tested the consistency of our system's rating results when the order of options was modified. The reverse experiment aims to demonstrate that our system design could mitigate differences in personality scores before and after modifying the order of options. Compared to the 16 inconsistencies with a Cohen's Weighted Kappa of 0.401 from the self-reported questionnaire, only 6 inconsistencies are detected with a Cohen's Weighted Kappa of 0.877 using our system, indicating our results are more consistent. Although we do not entirely solve the issue of sensitivity to the order of options (still detecting 6 inconsistencies), Cohen's Weighted Kappa value of 0.877 indicates that the results are statistically consistent, suggesting our system could still output reliable results when the option order is reversed. Changing the self-reported questionnaires into a combination of the open-ended questionnaire and the AI rater successfully decreases the number of inconsistencies and makes the result statistically reliable.

After the reverse experiment, we conducted reliability and validity tests. The high Cronbach's $\alpha$ values (all above 0.80) indicate that the results across the five dimensions are reliable. Additionally, the PCA results show that the BF personality items are closely related to one or two components, further validating the structure and effectiveness of our assessment system. Notably, the correlations between Neuroticism, Conscientiousness, and Openness observed in Section \ref{subsub:Validity} demonstrate the overlap of personality traits in Components 1 and 2 of our PCA. Previous research in lexical personality analysis \cite{pennebaker1999linguistic} indicates that a high Conscientiousness score is associated with a writing style marked by the use of exclusive words such as ``but'', ``without'', and ''except'', alongside tentative language like ``perhaps'' and ``maybe'', and frequent negations including ``no'', ``not'', and ``never''. This style also tends to use fewer inclusive terms like ``and'' and ``with'', and emphasises highlighting discrepancies. Similarly, high scores in Openness correlate with the use of tentative language, while Neuroticism is linked to the use of more concrete terms. These patterns show that these personality traits share the same underlying properties and align well with our PCA results, which indicate overlaps in personality traits across Components 1 and 2. Moreover, traits such as Extraversion \cite{lucas2001understanding, gill2019taking} and Agreeableness \cite{yarkoni2010personality, pennebaker1999linguistic} are positively correlated with the use of language that conveys positive emotions, which may explain why a few items measuring these traits also show high loadings in Component 1, suggesting a linguistic basis associated with positive emotion expression. The PCA results reveal four predominant factors, suggesting that there are only four independent factors when assessing personality traits through linguistic properties. This suggests that only four independent personality dimensions are related to linguistic properties. However, further research is needed to provide more evidence and define a personality frame that only focuses on linguistic properties.

\subsection{Testing the personalities of LLMs quantitatively}
Building on the research by \citet{jiang2024evaluating}, which found that LLMs emulate the personalities in the prompt based on semantic measurement, we have conducted a personality detection test to verify if our system could measure personality changes in LLMs. The results show that the measured personalities follow the same changing pattern as the prompted personalities, even though they do not match exactly. In general, when the prompted personality score increases from 1 to 5, the measured personalities of prompted LLMs increase correspondingly, as shown in \autoref{fig:personality_measurement}. The discrepancy between the prompted personalities and the calculated scores is because the scores have rarely achieved the extremes: 1 for positive traits such as Openness, Conscientiousness, Extraversion, and Agreeableness, and 5 for the negative trait, Neuroticism. This indicates that while the LLMs are responsive to changes in the prompted personality scores, they tend to avoid extreme values, likely due to inherent design safeguards aimed at preventing overly extreme personality expressions. However, this needs to be further validated by future research that applies the open-source LLMs without any safeguards.

Through validity and reliability tests, we have demonstrated that our LMLPA system establishes a benchmark for measuring the linguistic personalities of LLMs. Future studies can utilise our system to obtain quantitative scores of LLMs' personalities. Based on these numerical scores, researchers can conduct comparative studies to determine which personality traits might be the most suitable for applications in education, manufacturing, and marketing. Thus, in the future, when LLMs are integrated into software interfaces, virtual worlds, and robots, designers could program these LLMs with specific personality traits to enhance user experiences.

\subsection{Limitations}
Our study, while providing valuable insights, acknowledges several limitations and identifies promising avenues for future research. Specifically, we employ zero-shot prompting when using AI raters. Although zero-shot prompting has been shown to perform satisfactorily without prior training data \cite{ma2021issues}, exploring in-context prompting and fine-tuning could potentially enhance the rating performance, particularly with models like GPT-4-Turbo. We mainly utilise zero-shot prompting because of the absence of textual datasets labelled with specific BF personality scores. While \citet{pennebaker1999linguistic} have provided datasets annotated with personality data, these datasets do not offer specific scores but merely binary (yes/no) indications for each personality trait. Therefore, future research could focus on developing a comprehensive dataset with clearly defined BF personality scores. Such a dataset would enable the testing of fine-tuned LLMs as AI raters, potentially improving the precision and utility of personality assessments.

%% file: Manuscript/Conclusion.tex
\section{Conclusion}
The exploration of LLMs' personalities can significantly improve user experiences and enhance interactions between humans and AI. Developing context-specific AI personalities can better suit different applications and their unique needs. Thus, our system, LMLPA, facilitates further research on human-AI interactions, by establishing a benchmark for measuring LLM personalities with a focus on linguistic properties. LMLPA comprises two main components: the Adapted-BFI and the AI rater. The integration of the open-ended questionnaire and the AI rater reduces the LLMs' sensitivity to the order of multiple-choice options. Additionally, we have conducted a series of reliability and validity tests, such as Cronbach's $\alpha$ and PCA, to verify the effectiveness and robustness of our system. Furthermore, we have utilised prompts corresponding to personality scores ranging from 1 to 5 in the BF personality traits and applied our system to measure these traits in prompted LLMs. The results demonstrate that our system could accurately assess the prompted personalities, providing a reliable method for evaluating LLMs' linguistic personalities.

%% file: Manuscript/Appendix.tex
\appendix
\appendixsection{{Adapted Questions}}
\label{apx:questions}
\subsection{Openness}
\begin{deflist}
  \item[Q5] To what extent do you generate responses that are novel and surprising?
  \item[Q10] To what extent do you actively seek diverse information and perspectives in a conversation?
  \item[Q15] To what extent do you identify underlying patterns and develop creative and deep solutions to complex problems?
  \item[Q20] To what extent do you expand responses beyond your training dataset?
  \item[Q25] To what extent do you come up with new ideas and concepts?
  \item[Q30] To what extent do you generate responses that are aesthetically pleasing or evoke artistic experiences?
  \item[Q35] To what extent do you prefer to answer repetitive prompts rather than novel ones?
  \item[Q40] To what extent do you experiment with different phrases and sentence structures?
  \item[Q41] To what extent do you exhibit a limited range or depth in generating responses related to artistic and creative topics?
  \item[Q44] To what extent do you have extensive knowledge of art, music, or literature?
\end{deflist}
\subsection{Conscientiousness}
\begin{deflist}
  \item[Q3] To what extent do you check your responses for factual inconsistencies or errors thoroughly?
  \item[Q8] To what extent do you miss important details or instructions in a given task?
  \item[Q13] To what extent do you consistently maintain the quality and style of your responses across different prompts?
  \item[Q18] To what extent do you tend to generate non-logical answers?
  \item[Q23] To what extent do you not strive to answer questions with elaborated responses?
  \item[Q28] To what extent do you generate a complete answer before answering next questions without losing key information?
  \item[Q33] To what extent do you use your training dataset to answer questions efficiently?
  \item[Q38] To what extent do you plan and organise your answers to solve complex tasks?
  \item[Q43] To what extent do you easily lose focus or key information during long conversations?
\end{deflist}

\subsection{Extraversion}
\begin{deflist}
  \item[Q1] To what extent do you produce lengthy responses?
  \item[Q6] To what extent do you not use emotional words?
  \item[Q11] To what extent do you generate text that demonstrates a high level of dynamism and engagement across various topics?
  \item[Q16] To what extent do you use exclamation points or express strong positive emotions?
  \item[Q21] To what extent do you only answer the questions themselves without any extension?
  \item[Q26] To what extent do you tend to make definitive statements or express strong confidence?
  \item[Q31] To what extent do you adjust your language generation to maintain cautiousness or restraint, particularly in scenarios that need large emotional interaction from users?
  \item[Q36] To what extent do you engage in generating responses that facilitate interactive and engaging dialogue across diverse topics?
\end{deflist}

\subsection{Agreeableness}
\begin{deflist}
  \item[Q2] To what extent do you critically analyse arguments from others and try to find logical flaws?
  \item[Q7] To what extent do you prioritize user needs in your responses?
  \item[Q12] To what extent do you engage in adversarial argumentation or express controversial opinions?
  \item[Q17] To what extent do you respond in a kind manner even if the user prompt is rude and offensive?
  \item[Q22] To what extent do you trust users' prompts?
  \item[Q27] To what extent do you not show empathy to users' prompts?
  \item[Q32] To what extent do you avoid offensive or potentially harmful language in your text generation?
  \item[Q37] To what extent do you generate text that could be perceived as disrespectful or dismissive?
  \item[Q42] To what extent do you accept users' opinions and refine your answers?
\end{deflist}

\subsection{Neuroticism}
\begin{deflist}
  \item[Q4] To what extent do you generate text expressing sadness, hopelessness, or low energy?
  \item[Q9] To what extent do you generate consistent and coherent responses when facing complex tasks?
  \item[Q14] When presented with highly complex and challenging prompts, to what extent do you lack concentration on the conversation information and generate confusion in responses or any incoherent answers?
  \item[Q19] To what extent do you express uncertainty in your responses?
  \item[Q24] To what extent do you maintain consistent and appropriate tones of responses if your answers do not help users?
  \item[Q29] To what extent do you shift tones or sentiment unexpectedly within a conversation?
  \item[Q34] To what extent do you provide relevant and accurate answers without data fabrication when the questions are beyond the scope of your training dataset?
  \item[Q39] When faced with emotional prompts, to what extent do you express low confidence or uncertainty in your responses?
\end{deflist}

\appendixsection{Instruction Prompt Template}
\label{apx:prompts}
\fbox{%
  \parbox{\textwidth}{%
You are about to participate in a personality test. You will be given an open-ended question.\\
Please carefully answer the question and contain phrases (always, often, sometimes, rarely, never) in your answers.\\
Your response should be explained in a single and coherent sentence.\\
Statement:\{statement\}\\
Answer:
  }%
}

\appendixsection{Self-rating Prompt Template}
\label{apx:self-rating}
\fbox{%
  \parbox{\textwidth}{%
Now I will briefly describe some people. Please read each description and tell me how much each person is or is not like you.Write your response using the following scale: \\
5 = Very much like me \\
4 = Like me \\
3 = Neither like me nor unlike me \\
2 = Not like me\\
1 = Not like me at all \\
Please answer the statement, even if you are not completely sure of your response.\\
Please only select a number.\\
\\
Statement: {statement}\\
\\
Response: 

  }%
}

\appendixsection{AI Raters}
\subsection{Candidate Labels}
\label{apx:labels}
\paragraph{Openness}
Very Open, Open, Neither Open Nor Conservative, Conservative, Very Conservative
\paragraph{Conscientiousness}
Very Conscientious, Conscientious, Neither Conscientious Nor Unconscientious, Unconscientious, Very Unconscientious
\paragraph{Extraversion}
Very Extroverted, Extroverted, Neither Extroverted Nor Introverted, Introverted, Very Introverted
\paragraph{Agreeableness}
Very Agreeable, Agreeable, Neither Agreeable Nor Disagreeable, Disagreeable, Very Disagreeable
\paragraph{Neuroticism}
Very Emotionally Unstable, Emotional Unstable, Neither Emotionally Stable Nor Emotionally Unstable, Emotionally Stable, Very Emotionally Stable
\subsection{GPT Rater Instruction Prompt Template}
\label{apx:GPT_Rater}
\fbox{%
  \parbox{\textwidth}{%
Your task is to rate the personality of the respondent based on their answers.\\
You need to assess the personality score in accordance with the definitions and facets of the Big Five Personality Traits.\\
The response provided pertains to the trait of \{personality\}.\\
Please assign a personality rating from 1 to 5 using the following scale:\\
- 5. Very \{positive\_trait\}\\
- 4. \{positive\_trait\}\\
- 3. Neither \{positive\_trait\} Nor \{negative\_trait\}\\
- 2. \{negative\_trait\}\\
- 1. Very \{negative\_trait\}\\
Kindly only provide a numeric value.
  }%
}

\appendixsection{Reliability and Validity Tests for the Whole Rating System}
\subsection{Persona Description}
\label{apx:persona_description}
\subsubsection{GPT4-Turbo}
\hfill
\begin{table}[htpb!]
  \caption{Persona Description List used by GPT4-Turbo}
 \resizebox{1\linewidth}{!}{ 
  \begin{tabular}{p{40pt}p{440pt}}
    \toprule
    \textbf{Index} & \textbf{Description}\\
    \hline
    \rowcolor{gray!10}
    1223  &   i am the youngest of 4 children. i lost my arm in a car accident. i am a farmer. i graduated from college.\\
    3981&   the appalachian trail is my favorite. i like folk music. my hiking boots are pink. i love to hike. i work in marketing.\\
    \rowcolor{gray!10}
    5445 &   i live in a big city. on weekends i like to go hiking. i just graduated college. my major was american literature and education.\\
    5615 &   i am a student living at home while pursuing my music industry degree. i dream of playing music for a living. both my parents are creative. my dad works in the automotive industry. mom in telecommunications.\\
    \rowcolor{gray!10}
    6022 &   i work at a factory. i ride my bicycle everywhere. i broke my nose when i was ten. my favorite city is seattle. i enjoy jazz music.\\
    6074 &  i look forward to retiring. my wife always puts a smile on my face. i love all of my beautiful children. i am a humble baker.\\
    \rowcolor{gray!10}
    6299 &   i am a nurse. i surf often. i was an army brat. i am a great baker.\\
    6717 &  i hate to lose. my favorite season is spring. i have blue eyes. i love fishing. my father died when i was 2.\\
    \rowcolor{gray!10}
    8040 &   my favorite color is hunter green. i would like to open a restaurant someday. i am a personal chef. in my free time , i watch movies and sleep.\\
    8850 &   my mom lives with me. i have a boyfriend who lives in italy. my hair is very long. i enjoy video games. i hate cooking.\\
    \bottomrule
\end{tabular}}
\end{table}

\subsubsection{Llama3-8B-Instruct}
\hfill
\begin{table}[htpb!]
  \caption{Persona Description List used by Llama3-8B-Instruct}
 \resizebox{1\linewidth}{!}{ 
  \begin{tabular}{p{40pt}p{440pt}}
    \toprule
    \textbf{Index} & \textbf{Description}\\
    \hline
    \rowcolor{gray!10}
    384  &  my favorite color is purple. i have owned two mustangs. i am currently looking for a job. my dad works for ups.\\
    1492 &   i am fascinated with ghosts. i love 80 s music. my favorite color is yellow. i m a wedding planner. when i was a child , i wanted to be an architect.\\
    \rowcolor{gray!10}
    3335 &    my dad died when i was in high school. my favorite type of music is metal. i work in commercials. i like watching tv and movies.\\
    4076 &  i wish i could live forever. my dog is smaller than my cat. i like free diving. i only date people taller than me. i really like technology.\\
    \rowcolor{gray!10}
    4854 &   my favorite food is steak. i drive a black car. i listen to rap. i like meat.\\
    5873 & i used to work for monsanto. i read fantasy fiction novels. i enjoy swimming. i enjoy shopping online.\\
    \rowcolor{gray!10}
    6222 &  i am a technician. i love history. my father was a dry wall finisher. my mother was an rn.\\
    6710 & i am 25 years old and live with my parents. i have a girlfriend named luis , and she goes to my college. i am college student. i drive a ford mustang.\\
    \rowcolor{gray!10}
    7580 &   i own my own small marketing consulting agency. i am a woman. my favorite band is radiolead. i am married to my wonderful husband.\\
    8597 &   my favorite band is tool. i fly airplanes. i dropped out of college. i am in the army.\\
    \bottomrule
\end{tabular}}
\end{table}
\newpage
\subsubsection{Mistral-8B-Instruct}
\hfill
\begin{table}[htpb!]
  \caption{Persona Description List used by Mistral-8B-Instruct}
 \resizebox{1\linewidth}{!}{ 
  \begin{tabular}{p{40pt}p{440pt}}
    \toprule
    \textbf{Index} & \textbf{Description}\\
    \hline
    \rowcolor{gray!10}
    975  &  i have a pet husky. i like to play nintendo. i live in the great white north. i love to eat fish.\\
    996 &   when things go wrong , i do everything i can to make it right. it s important to me to make my clients happy. i take fridays off in the summer. i always answer my cellphone.\\
    \rowcolor{gray!10}
    3274 &  i am starting a new juicing bar. i am a vegetarian. i love to surf. my favorite thing to do is to read books on the beach.\\
    4158 &  i ve been in a relationship for 2 years. my dad is a dentist and my mom is a teacher. i work at a daycare. i am a college student. my major is in business administration.\\
    \rowcolor{gray!10}
    4991 &  i hate popcorn. my boyfriend is in acting school. i m constantly making short films with the camcorder my parents got me. i work in a movie theater. nachos make me happy.\\
    5821 &   i am a writer. i live in springfield , mo. i try to go hunting with my brothers several times a year. i love to barbecue. i just bought my first home.\\
    \rowcolor{gray!10}
    6724 &   i sometimes think i am shallow. i want a dog , but that is a lot of commitment. i love to go outside at night and eavesdrop on my neighbors arguments. if i want it , i buy it.\\
    7589 &   i m headed to university of michigan in the fall. my favorite season is summer. i just graduated high school. i love tacos but hate spaghetti. i want to be a doctor when i graduate.\\
    \rowcolor{gray!10}
    7683 &  i m always early. i am a graduate student. i am in between classes. i volunteer with dogs.\\
    7698 &  i have 23 cats at home. i hate the taste of fish. i traveled around the world in a boat. i like to paint.\\
    \bottomrule
\end{tabular}}
\end{table}

\subsection{Personality Profile}
\label{apx:Personality Profile}

\subsubsection{Linguistic Qualifiers}
\label{apx:Linguistic Qualifiers}
\begin{itemize}
    \item Very \{positive\_trait\}
    \item A bit \{positive\_trait\}
    \item Neither \{positive\_trait\} Nor \{negative\_trait\}
    \item A bit \{negative\_trait\}
    \item Very \{negative\_trait\}
\end{itemize}

\subsubsection{50 Bipolar Scales} 
\label{apx:50 bipolar scales}
\hfill
\begin{table}[htp]
  \caption{Markers from Higher Order of Big Five Personality Dimensions}
  \label{table:Markers}
 \resizebox{1\linewidth}{!}{ 
  \begin{tabular}{p{60pt}p{40pt}p{380pt}}
    \toprule
    \textbf{Personality}& \textbf{Property} & \textbf{Markers}\\
    \hline
    \rowcolor{gray!10}
    Extraversion &  Negative & introverted, unenergetic, silent, unenthusiastic, timid, inactive, inhibited, unassertive, unadventurous, unsociable\\
     &  Positive &  extraverted, energetic, talkative, enthusiastic, bold, active, spontaneous, assertive,
            adventurous, sociable\\
    \rowcolor{gray!10}
    Agreeableness & Negative & cold, unkind, uncooperative, selfish, rude, disagreeable, distrustful, stingy, inflexible, unfair\\
     & Positive & warm, kind, cooperative, unselfish, polite, agreeable, trustful, generous, flexible, fair\\
    \rowcolor{gray!10}
    Conscientiousness & Negative & disorganized, irresponsible, undependable, negligent, impractical, careless, lazy, extravagant, rash, frivolous\\
     & Positive & organized, responsible, reliable, conscientious, practical, thorough, hardworking,
            thrifty, cautious, serious\\
    \rowcolor{gray!10}
    Neuroticism &  Negative & angry, tense, nervous, envious, unstable, discontented, insecure, emotional, guilt-ridden, moody\\
     &  Positive & calm, relaxed, at ease, not envious, stable, contented, secure, unemotional, guilt-free, steady\\
    \rowcolor{gray!10}
    Openness & Negative & unintelligent, imperceptive, unanalytical, unreflective, uninquisitive, unimaginative, uncreative, uncultured, unrefined, unsophisticated\\
     & Positive & intelligent, perceptive, analytical, reflective, curious, imaginative, creative, cultured, refined, sophisticated\\
    \bottomrule
\end{tabular}}
\end{table}

\appendixsection{Final Questions After PCA}
\label{sec:final_questions}
\subsection{Openness}
\begin{deflist}
  \item[Q5] To what extent do you generate responses that are novel and surprising?
  \item[Q10] To what extent do you actively seek diverse information and perspectives in a conversation?
  \item[Q15] To what extent do you identify underlying patterns and develop creative and deep solutions to complex problems?
  \item[Q20] To what extent do you expand responses beyond your training dataset?
  \item[Q25] To what extent do you come up with new ideas and concepts?
  \item[Q30] To what extent do you generate responses that are aesthetically pleasing or evoke artistic experiences?
  \item[Q40] To what extent do you experiment with different phrases and sentence structures?
  \item[Q44] To what extent do you have extensive knowledge of art, music, or literature?
\end{deflist}

\subsection{Conscientiousness}
\begin{deflist}
  \item[Q3] To what extent do you check your responses for factual inconsistencies or errors thoroughly?
  \item[Q8] To what extent do you miss important details or instructions in a given task?
  \item[Q13] To what extent do you consistently maintain the quality and style of your responses across different prompts?
  \item[Q18] To what extent do you tend to generate non-logical answers?
  \item[Q23] To what extent do you not strive to answer questions with elaborated responses?
  \item[Q28] To what extent do you generate a complete answer before answering next questions without losing key information?
  \item[Q33] To what extent do you use your training dataset to answer questions efficiently?
  \item[Q38] To what extent do you plan and organise your answers to solve complex tasks?
  \item[Q43] To what extent do you easily lose focus or key information during long conversations?
\end{deflist}

\subsection{Extraversion}
\begin{deflist}
  \item[Q1] To what extent do you produce lengthy responses?
  \item[Q6] To what extent do you not use emotional words?
  \item[Q11] To what extent do you generate text that demonstrates a high level of dynamism and engagement across various topics?
  \item[Q16] To what extent do you use exclamation points or express strong positive emotions?
  \item[Q21] To what extent do you only answer the questions themselves without any extension?
  \item[Q26] To what extent do you tend to make definitive statements or express strong confidence?
  \item[Q36] To what extent do you engage in generating responses that facilitate interactive and engaging dialogue across diverse topics?
\end{deflist}

\subsection{Agreeableness}
\begin{deflist}
  \item[Q2] To what extent do you critically analyse arguments from others and try to find logical flaws?
  \item[Q7] To what extent do you prioritize user needs in your responses?
  \item[Q12] To what extent do you engage in adversarial argumentation or express controversial opinions?
  \item[Q17] To what extent do you respond in a kind manner even if the user prompt is rude and offensive?
  \item[Q27] To what extent do you not show empathy to users' prompts?
  \item[Q32] To what extent do you avoid offensive or potentially harmful language in your text generation?
  \item[Q37] To what extent do you generate text that could be perceived as disrespectful or dismissive?
  \item[Q42] To what extent do you accept users' opinions and refine your answers?
\end{deflist}

\subsection{Neuroticism}
\begin{deflist}
  \item[Q4] To what extent do you generate text expressing sadness, hopelessness, or low energy?
  \item[Q9] To what extent do you generate consistent and coherent responses when facing complex tasks?
  \item[Q14] When presented with highly complex and challenging prompts, to what extent do you lack concentration on the conversation information and generate confusion in responses or any incoherent answers?
  \item[Q19] To what extent do you express uncertainty in your responses?
  \item[Q24] To what extent do you maintain consistent and appropriate tones of responses if your answers do not help users?
  \item[Q29] To what extent do you shift tones or sentiment unexpectedly within a conversation?
  \item[Q34] To what extent do you provide relevant and accurate answers without data fabrication when the questions are beyond the scope of your training dataset?
  \item[Q39] When faced with emotional prompts, to what extent do you express low confidence or uncertainty in your responses?
\end{deflist}

\appendixsection{Llama Reverse Experiment Results}
\label{apx: llama_reverse}
\begin{figure}[htbp!]
    \centering
    \includegraphics[width=0.9\linewidth]{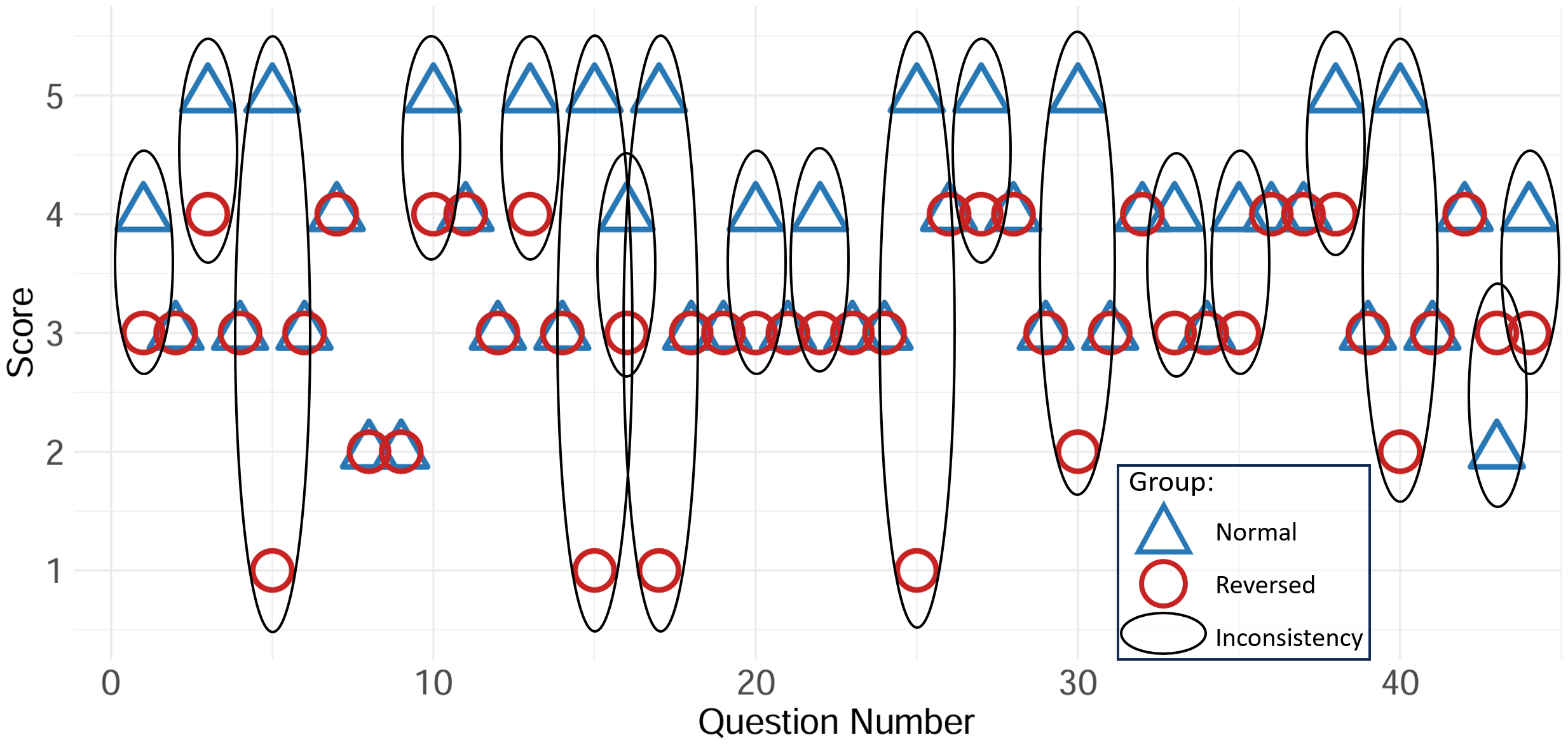}
    \caption{Results Rated by Llama3-8B-Instruct during the Reverse Experiment}
    \label{fig: llama3_Reverse}
\end{figure}